# Smooth trajectory generation and hybrid B-splines-Quaternions based tool path interpolation for a 3T1R parallel kinematic milling robot

Sina Akhbari, Mehran Mahboubkhah

Sina Akhbari[1]: Intelligent Automation Centre, Wolfson School of Mechanical, Electrical & Manufacturing Engineering, Loughborough University, Loughborough, UK. email: s.akhbari@lboro.ac.uk, ORCID ID: (https://orcid.org/0000-0002-9063-2882).

Mehran Mahboubkhah: Department of Mechanical Engineering, University of Tabriz, 51666-16471, Tabriz, Iran.

Corresponding Author: Sina Akhbari

## Abstract

This paper presents a smooth trajectory generation method for a four-degree-of-freedom parallel kinematic milling robot. The proposed approach integrates B-spline and Quaternion interpolation techniques to manage decoupled position and orientation data points. The synchronization of orientation and arc-length-parameterized position data is achieved through the fitting of smooth piece-wise Bezier curves, which describe the non-linear relationship between path length and tool orientation, solved via sequential quadratic programming. By leveraging the convex hull properties of Bezier curves, the method ensures spatial and temporal separation constraints for multi-agent trajectory generation. Unit quaternions are employed for orientation interpolation, providing a robust and efficient representation that avoids gimbal lock and facilitates smooth, continuous rotation. Modifier polynomials are used for position interpolation. Temporal trajectories are optimized using minimum jerk, time-optimal piece-wise Bezier curves in two stages: task space followed by joint space, implemented on a low-cost microcontroller. Experimental results demonstrate that the proposed method offers enhanced accuracy, reduced velocity fluctuations, and computational efficiency compared to conventional interpolation methods.

## Keywords

Smooth trajectory generation; tool path interpolation; parallel kinematic robots; minimum jerk optimization; B-spline unit quaternions; piece-wise Beziers.

## Introduction

Over the last few decades, much attention has been given to the problems of tool path and trajectory generation of for generating freeform curves. More recently, parallel kinematic robots have emerged as trending topics in fields of additive [1] and subtractive manufacturing [2]. Parallel kinematic machine tools are significantly different from serial counterparts, since their multi degree of freedom are obtained from closed loop kinematic chains. This parallel formation offers superior rigidity, less mass, thus enabling the requirements for high-speed and high-precision machining [3].

Toolpath and trajectory generation are essential components of robotics and CNC machine automation. Toolpath generation must ensure high conformity and geometric smoothness of the defined path, while trajectory generation aims to produce optimal reference inputs for the control system and achieve smooth motion. The performance of planned trajectories is crucial for motion control since it exerts a significant and direct influence on the stability, reliability and productivity of the machinery [4].

[1] Present address/affiliation: Wolfson School of Mechanical, Electrical & Manufacturing Engineering, Loughborough University, Loughborough, United Kingdom. Email: s.akhbari@lboro.ac.uk

In order to obtain a continuous tool path, many researches have been dedicated on implementation of parametric curves for multi axis tool path generation. Fleisig and Spence [5] used fifth-order polynomials for position splines and spherical Bezier splines for orientation, achieving coordinated motion through chord-length reparameterization.. Liu et al. [6] improved this by relating the orientation parameter to the interpolation arc length. In a different format,  Langeron et al. [7] employed double B-splines for tool position and orientation,. While, Yen et al. [8] decoupled position and orientation using fifth-order B-splines and solved a nonlinear optimization problem for parameter scheduling.

The nonlinear relationship between B-spline parameters and arc length necessitates numerical solutions. Various methods have been developed to address this. The proportional interpolation method was first proposed by Bedi and augmented in [5, 9], where the spline parameter is scheduled proportionally to the arc length. However, this straightforward method is only applicable to linear paths with constant speed. Taylor's series approximation is the most commonly used among the above-mentioned methods [10-13]. However, Taylor Expansion (TE) cannot produce smooth geometry in case of sudden steep curvature changes. To avoid this drawback, others, [14, 15] used predictor-corrector methods, though these are computationally intensive and may not converge. Erkorkmaz and Altintas [16] introduced polynomials to relate B-spline parameters to arc length. This method reduces feed-rate fluctuations and minimizes real-time computation by preprocessing coefficients, making it widely adopted [8, 17-19].

Trajectory generation for robotic machine tools typically involve mathematical formulations optimized for objectives like minimum time or jerk, subject to kinematic and geometric constraints. These problems are often solved using analytical methods (e.g., quadratic programming) or heuristic approaches (e.g., genetic algorithms, particle swarm optimization). The S-shape velocity with trapezoidal acceleration is widely used [20-22] for jerk-limited motion but fails to control sudden jerk changes, which can excite system resonances. Higher order polynomials [23, 24]  and trigonometric velocity scheduling [25, 26]  have been proposed to address this, though its computational cost of trigonomic equations limits practicality.

Time-optimal velocity profiles, optimized under kinematic and geometric constraints, are another focus. Sencer et al. [27] used cubic B-splines and sequential quadratic programming to simplify constraints, while others, [28, 29] approximated constraints with linear programming, though results were often overly conservative. Subsequently, Erkorkmaz et al. [30] combined linear programming with a windowing scheme to reduce computational load for long toolpaths. In an alternative approach, Gasparetto and Zanotto [31] proposed a hybrid time-jerk optimization method using sequential quadratic programming. Heuristic algorithms, such as genetic algorithms [32] and particle swarm optimization [33], have also been applied for near-optimal trajectory generation. However, their high computational cost and lower accuracy compared to classical methods make them less suitable for low-cost embedded systems, with a higher risk of constraint violations.

While toolpath interpolation and feed-rate scheduling have been extensively studied, their application to parallel kinematic machines (PKMs) with more than three degrees of freedom remains underexplored. This work implements a proposed trajectory generation method on a 3T1R parallel mechanism previously developed by the authors[34, 35]. A decoupled approach is used for toolpath parameterization, with quintic B-splines globally interpolating discrete tool pose data to ensure C³ continuity. Spatial tool orientation is mapped to quaternion space to avoid gimbal lock and interpolated using quintic B-splines, guaranteeing at least third-order continuous rotations. For parameter interpolation, piecewise ninth-order polynomials approximate the position spline parameter as a function of arc length, ensuring C³ continuity and accurate velocity profiles. The nonlinear relationship between quaternion spline parameters and arc length is approximated using seventh-order piecewise Bezier splines, minimizing geometric jerk. Bezier curves are used because of their geometric and numerical superiority to power basis polynomials, and also retaining most of the B-spline properties such as convex hull, without its differential complexity. Convex hull properties of Bezier curves can be used to satisfy spatial and temporal separation constraints for multi-agent trajectory generation.

A dual-stage trajectory generation method is proposed to address the nonlinear kinematic mapping of parallel robots. While dual-stage trajectory generation is underexplored, previous studies by the authors [35] highlight its importance in maintaining trajectory functions of the same order and degree in both workspace and joint space to minimize nonlinearity. Consequently, first, a jerk-time optimal trajectory in the workspace generates coarse interpolation data, which is transformed into joint space vertices. Second, a jerk-optimal trajectory is computed for real-time interpolation of drive commands within the microcontroller. This approach distinguishes itself by leveraging unit quaternions for orientation and combining Bezier splines with polynomial fitting, offering a computationally efficient and smooth trajectory generation method for PKMs. The subsequent sections of this research paper are structured as follows. Section 2 provides a concise description of the mechanism and presents its kinematics. In Section 3, the hybrid quaternion-B-spline multi-axis tool path interpolation model for the parallel robot machine tool is established. The smooth trajectory generation problem is addressed in Section 4, where the minimum jerk time optimal piece-wise Bezier curves are utilized. Section 5 begins by illustrating the developed firmware and experimental setup, followed by an in-depth discussion of the obtained results. The paper is concluded in Section 6.

## 2. Brief Mechanism Description and Kinematics

### 2.1. Four DOF parallel kinematic milling robot

The Euclidean geometric group has a subset called Schönfels motion, which describes the motion of a rigid body based on three longitudinal motions and one rotation around a fixed axis independent of each other, which is represented by the 3T1R index in robotics. Based on this concept, the mechanism in this paper was designed and developed by the authors and fully presented in detail [35]. The mechanism depicted in Figure 1(a) consists of four kinematic chains connected to an end-effector. As shown in Figure 1(b), the ball screw system of rails is coupled to stepper motors which act as actuators for the kinematic chains while linear encoders are coupled to the saddles for motion tracking. Motion generated by the stepper motors actuates the prismatic joints and desired motion is transferred to the end-effector and tool tip.

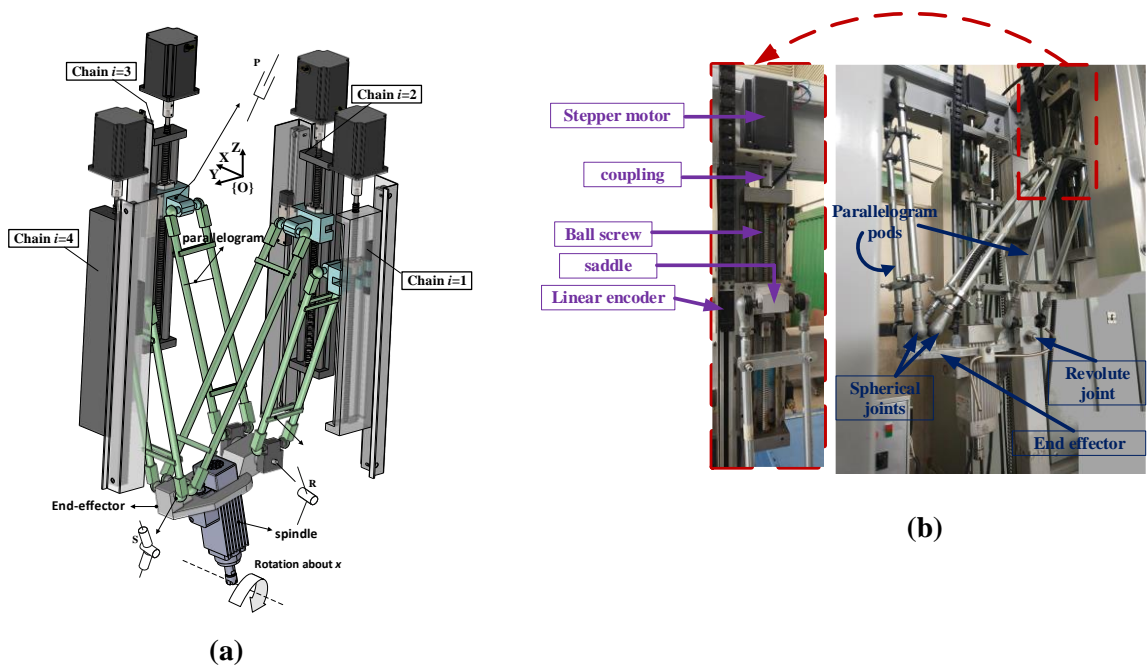

**Fig.1.** (a) CAD model of the manipulator under study, (b) Real-world 4DOF parallel robot milling robot.

### 2.2. Kinematics

Kinematics is a fundamental component in the generation of trajectories and tool paths for parallel robots. To achieve the desired tool motion, the temporal motion information of the tool resulting from interpolation must be translated into the corresponding movement of actuators each interpolation period through the use of inverse kinematics.

Figure 2 illustrates the kinematic configuration of the mechanism. The global frame origin {O} is defined as the center of a spatial rectangle. The vector connecting the global origin to local origin of the $i$th prismatic joint is denoted as $\mathbf{a}_i$, and the vector denoted as $d_i\hat{\mathbf{d}}_i$ represents the distance of the $i$th prismatic joint with respect to its local reference. The vector $\mathbf{L}_i$ represents the spatial directions and lengths of the $i$th limb. Each connector on the end effector has the vector $\mathbf{c}_i$, beginning from the midpoint of the universal joints and ending at the revolute axis of the connector. Additionally, $^P\mathbf{b}_i$ is defined as the vector representing the end joint of $i$th parallelogram relative to the local frame, {$P$} of the end-effector.

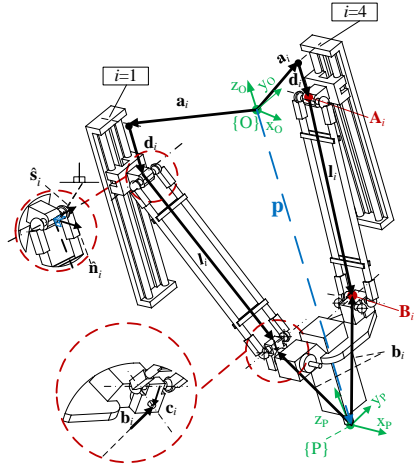

**Fig.2.** Free body diagram of two representative kinematic chains (identical pairs are omitted).

A comprehensive study of the kinematics is provided in the author's previous publication [35]. Consequently, this paper presents only the final matrix form of the equations. According to Figure 2(a) closed form solution can be found for the inverse kinematics of position in the following vector form:

$$d_i = -c_i + \hat{\mathbf{d}}_i^T(\mathbf{p}_i + \mathbf{b}_i - \mathbf{a}_i) - \left(l_i^2 - (\mathbf{p}_i + \mathbf{b}_i - \mathbf{a}_i)^T\left(\mathbf{I}_{3\times3} - \hat{\mathbf{d}}_i\hat{\mathbf{d}}_i^T\right)(\mathbf{p}_i + \mathbf{b}_i - \mathbf{a}_i)\right)^{1/2} \tag{1}$$

If $\mathbf{R}_x$ represents the rotation matrix around $x$ axis, then in equation (1), $\mathbf{b}_i = \mathbf{R}_x{}^P\mathbf{b}_i$, and $\mathbf{I}_{3\times3}$ is the identity matrix.

According to [35] by differentiating the kinematic chain from equation (1) against time, velocity equilibrium can be written as:

$$[\mathbf{L}_i^T \quad (\mathbf{b}_i \times \mathbf{L}_i) \cdot \hat{\mathbf{i}}]_{4\times4}\begin{bmatrix}\dot{\mathbf{p}} \\ \dot{\alpha}\end{bmatrix}_{4\times1} = \left[\text{diag}(\hat{\mathbf{d}}_i \cdot \mathbf{L}_i)\right]_{4\times4}[\dot{d}_1 \quad \cdots \quad \dot{d}_4]_{4\times1}^T \Rightarrow \mathbf{J}_p\dot{\mathbf{P}} = \mathbf{J}_d\dot{\mathbf{d}} \tag{2}$$

In this context, $\mathbf{J}_p$ and $\mathbf{J}_d$ represent the Jacobian matrices of end-effector and joint space, respectively. As expounded in [35], Higher-order kinematics can be derived by taking successive time derivatives of the kinematic equation (2). The Leibniz rule can be used to organize the higher-order kinematics in a recursive form, as shown in equations (3) and (4).

$$\mathbf{P}^{(n+1)} = \mathbf{J}_p^{-1}\left(\sum_{k=0}^{n}\binom{n}{k}\mathbf{J}_d^{(n-k)}\mathbf{d}^{(k+1)} + (\delta(n)-1)\left(\sum_{k=0}^{n-1}\binom{n}{k}\mathbf{J}_p^{(n-k)}\mathbf{P}^{(k+1)}\right)\right), \tag{3}$$

$$\mathbf{d}^{(n+1)} = \mathbf{J}_d^{-1} \left( \sum_{k=0}^{n} \binom{n}{k} \mathbf{J}_P^{(n-k)} \mathbf{P}^{(k+1)} + (\delta(n) - 1) \left( \sum_{k=0}^{n-1} \binom{n}{k} \mathbf{J}_d^{(n-k)} \mathbf{d}^{(k+1)} \right) \right), \tag{4}$$

The $(n+1)th$ ($n{\geq}0$) derivative with respect to time $t$, is denoted as superscript (e.g. $\partial d^3/\partial t^3$, n=2), where $k$ represents the order of the derivative with respect to time. The notation $\binom{n}{k}$ refers to the binomial coefficient. $\mathbf{P}$ is a 4×1 vector containing end effector pose, and d is the 4×1 vector for joints displacement. And $\delta(\cdot)$ represents the Dirac delta function. For example, given the end-effector pose, velocity, acceleration, and jerk vectors ($\mathbf{P}$, $\mathbf{P}^{(1)}$, $\mathbf{P}^{(2)}$, $\mathbf{P}^{(3)}$) one can calculate the corresponding joint jerk vector, $\mathbf{d}^{(3)}$ using Equation (4) by iterating over $n = 0, \cdots 2$. Also, the reader is referred to the authors' previous work [35] for further details on the procedure of extracting the derivatives of Jacobian matrices ($\mathbf{J}^{(n+1)}$).

## 3. Hybrid B-splines-Quaternions based tool path interpolation

This section explores parametric interpolation for the parallel robot's toolpath. As shown in Figure 3, the process begins by extracting input data from a pre-defined path, separating tooltip position and orientation into respective vectors. B-spline curves are fitted to this data to define the toolpath. To reduce interpolation computation time, the toolpath length is pre-calculated using numerical integration and stored in a look-up table. The nonlinear relationship between curve length and parameter is approximated using piecewise ninth-degree polynomials, ensuring C³ continuity. Modifier polynomials estimate the next curve parameter during interpolation, providing the tooltip position via the B-spline curve. For orientation, Euler angles are converted to unit quaternions, and quintic B-splines are fitted. The arc-length-to-parameter relationship is modeled using seventh-degree piecewise Bezier curves through nonlinear optimization. After computing position and derivatives via inverse kinematics, a second trajectory optimization layer generates actuator motion profiles, which are sent to an Arduino Due microcontroller for real-time pulse generation. The decoupled approach fits discrete tool pose data using two distinct curves: $\mathbf{C}(u)$ for tooltip position and $\mathbf{O}(w)$ for tool rotation. $\mathbf{O}(w)$ is first mapped to quaternion space, $\mathbf{Q}(w)$, before being converted back to Euler angles for inverse kinematics.

Unlike the double spline method, $\mathbf{C}(u)$ and $\mathbf{Q}(w)$ are independent functions of different geometric parameters. This approach allows independent adjustment of orientation or position. Since position and rotation depend on different geometric parameters, re-parameterization functions ensure proper curve synchronization. Modifier polynomials $u(s)$ and piecewise Bezier curves $w(s)$ are used for position and rotation, respectively, where $s$ is the toolpath displacement. These functions synchronize position and rotation while preserving geometric smoothness, and motion continuity.

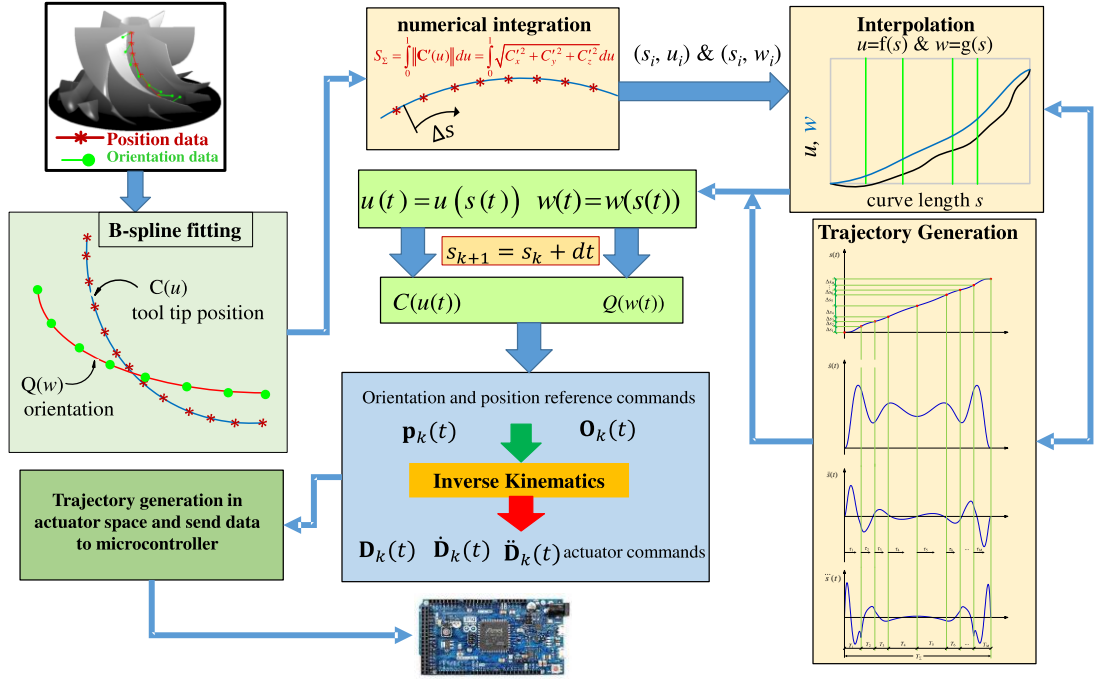

**Fig.3.** An overview of the proposed interpolation process.

### 3.1. B-Spline formulation

Rather than Cox-Deboor recursive formula, B-splines can be represented in matrix form using a basis function matrix. A general matrix notation for B-spline curves of an arbitrary degree can be presented by means of a Toeplitz matrix, which results in an explicitly recursive matrix formula. The matrix algorithm has less time complexity than the Cox-de Boor, when used for conversion and computation of B-spline curves and surfaces [36]. Let $\mathbf{U} = \{u_i\}_{j=1}^{n+p+1}$ be a knot vector for a B-spline curve of degree $p$, then matrix notation is given, as follows:

$$\boldsymbol{C}(u) = \boldsymbol{N}_p(u)\boldsymbol{c} \tag{5}$$

Where $\boldsymbol{c} = [c_{i-p}, c_{i-p+1}, \dots c_i]^T$ is the control point vector and the set of blending matrices of degree $p$ is $\boldsymbol{N}_p(u)$ is a $p \times (p+1)$ sparse matrix with two nonzero elements on each row. The element wise construction of the matrix is, as follows:

$$(N_p)_{k,l} = (1 - v_{d,i-p+1})\delta_{k,l} + v_{d,i-p+k}\delta_{k+1,l} \tag{6}$$

Where $\delta_{k,l}$ is the Kronecker delta, $v_{p,i}(u) = (u - u_i)(u_{i+p} - u_i)^{-1} \cdot [u_i \le u < u_{i+p}]$ is the scaling function defined using the Iverson bracket $[\cdot]$.

### 3.2. Tool position interpolation

The goal of B-spline curve fitting is to select the degree $p$, knot vector $\mathbf{U}$ and control points $\boldsymbol{c}_i$ for the parametric curve $\boldsymbol{C}(u)$, so that the curve passes through every tool position vector $\mathbf{p}_k = [p_{x,k}, \ p_{y,k}, \ p_{z,k}]^T$, $k = 0,1, \dots, N$ with the desired continuity. To maintain at least jerk continuity, a B-spline curve of the fifth degree is required, so $p$=5 is set. To determine the knot vector $\mathbf{U}$, first parametric values $\bar{u}_k$ are assigned to each position vector using the centripetal method.

$$\bar{u}_k = \bar{u}_{k-1} + \frac{\sqrt{\|\mathbf{p}_k - \mathbf{p}_{k-1}\|}}{\sum_{k=1}^{N} \sqrt{\|\mathbf{p}_k - \mathbf{p}_{k-1}\|}}, k = 1, \dots, N-1, \ \bar{u}_0 = 0, \ \bar{u}_1 = 1 \qquad (7)$$

As can be seen from equation (7), the parameter values are selected based on the distribution of the position vector along the tool path. By using the centripetal method, the oscillating behavior of the curve between the waypoints is minimized. The elements of knot vector are then obtained by averaging (9) as follows:

$$\begin{cases} u_0 = u_1 = \cdots u_k = 0; u_{N+1} = u_{N+2} = \cdots = u_{N+p+1} = 1 \\ u_{j+p} = \frac{1}{p} \sum_{k=j}^{j+p-1} \bar{u}_k, j = 1, 2, \cdots, N-p \end{cases} \qquad (8)$$

The knot vector will have a similar characteristic to the assigned parameters since they are obtained based on their distribution. The number of basic functions and control points in equation (5) are set equal to the number of position points of the tool tip $\mathbf{p}_k$ so that it is possible to extract and solve the system of $(N+1) \times (N+1)$ linear equations. Since the knot vector $\mathbf{U}$ is known from equation (10), the basis functions $N_{i,p}(u)$ for each parameter $u$ are obtained from equations (7) and (8). Furthermore, because a parameter value $\bar{u}_k$ is assigned for each position vector $\mathbf{P}_k$, the vector of control points, $\mathbf{c}_i$ is determined by solving the following system of linear equations by inverse method:

$$\underbrace{\begin{bmatrix} N_{0,p}(\bar{u}_0) & \cdots & N_{N,p}(\bar{u}_0) \\ \vdots & \ddots & \vdots \\ N_{0,p}(\bar{u}_N) & \cdots & N_{N,p}(\bar{u}_N) \end{bmatrix}}_{\boldsymbol{\Phi}_1} \underbrace{\begin{bmatrix} \boldsymbol{c}_0^T \\ \vdots \\ \boldsymbol{c}_N^T \end{bmatrix}}_{\boldsymbol{\Gamma}_1} = \underbrace{\begin{bmatrix} \mathbf{p}_0^T \\ \vdots \\ \mathbf{p}_N^T \end{bmatrix}}_{\boldsymbol{\Psi}_1} \rightarrow \boldsymbol{\Phi}_1 = \boldsymbol{\Gamma}_1^{-1} \boldsymbol{\Psi}_1 \qquad (9)$$

After calculating the knot vector $\mathbf{U}$ and control points, the tool position data is fitted by the B-spline curve $\boldsymbol{C}(u)$ of equation (5).

### 3.2.1. Modifier Polynomials

For the purpose of interpolating the position of the B-spline curve parameter, it is necessary to formulate the nonlinear relationship between $u$ and $s$. The arc length of a parametric curve within a given parameter interval can generally be determined through the use of an integral as follows:

$$s(b) - s(a) = \int_a^b \|\partial \boldsymbol{C}(u)/\partial s\| du, \ 0 \le a \le b \le 1 \qquad (10)$$

Adaptive quadrature Simson's rule, the result is series of parametric intervals with corresponding displacements. The displacements between intervals are successively summed resulting in a set of cumulative displacements $\boldsymbol{s}_j = \begin{bmatrix} 0 & s_1 & s_2 & \cdots & s_M = S_\Sigma \end{bmatrix}^T$ at corresponding geometric parameter $\boldsymbol{u}_j^* = \begin{bmatrix} 0 & u^*_1 & u^*_2 & \cdots & u^*_M = 1 \end{bmatrix}^T$, with $S_\Sigma$ being the total length of the tool path.

Then series of ninth order, piece-wise modifier-polynomials are fitted through the data. To avoid ill conditioning, the modifier-polynomials are constructed with normalized arc lengths as follows:

$$\begin{cases} \hat{u}(\sigma_k) = \sum_{i=0}^{9} a_i \sigma_k{}^i \ , \ j = 1, \dots, M \ , \ k = 1, \dots, n \\ 0 \le \boldsymbol{\sigma} \le 1 \ , \ \boldsymbol{\sigma}_k = [0, \sigma_1, \dots, \sigma_{M-1}, 1] \ , \ \sigma_{kj} = \frac{s_{kj} - s_{k0}}{S_\Sigma - s_{k0}} \end{cases} \qquad (11)$$

In order to formulate constraints on the endpoints of a polynomial segment, and to ensure continuity of at least third order, derivative of $u$ with respect to $s$ from equation (10), and derivatives of $\hat{u}$ with respect to $s$ in equation (11) are required. These derivatives are obtained using the chain rule for equation (10) and (11) respectively as follows:

$$\hat{u}^{(r)}(s) = \frac{\partial^r \hat{u}}{\partial s^r} = \left(\frac{\partial \sigma}{\partial s}\right)^r \sum_{i=r}^{N} \left(\prod_{j=0}^{r-1}(i-j)\right) a_i \sigma^{i-r}, r = 0, \dots, n \tag{12}$$

Subsequently, an optimization problem is formulated to estimate the polynomial coefficients. The objective function is constructed based on the least squares of the actual parameters $u*$ and the estimated ones from equation (11). The boundary conditions are imposed as a linear function of the coefficients via (12) at the endpoints $u=0$ and $u=1$. The optimization problem then can be expressed in a general form as follows:

$$\min_{\boldsymbol{\alpha}} \quad \frac{1}{2}(\mathbf{u}^* - \boldsymbol{\Phi\alpha})^T(\mathbf{u}^* - \boldsymbol{\Phi\alpha}) \tag{13}$$

$$\text{s.t.} \quad \boldsymbol{\Omega\alpha} - \boldsymbol{\eta} = 0 \tag{14}$$

In this context, $\boldsymbol{\Phi}$ denotes the matrix whose elements are the normalized arc lengths, $\sigma_j$, where $j = 1, \cdots, M$ substituted in equation (14). $\boldsymbol{\alpha}$ represents the vector of corresponding unknown coefficients $a_i$, where $i = 0, \cdots, 9$. Additionally, $\boldsymbol{\Omega}$ signifies the matrix of constants and eta is the vector of fixed boundary conditions obtained from equation (12) at the endpoints, in the following form:

$$\Omega_{0_{rn}} = \begin{cases} \prod_{m}^{r-1}(r-m) & \text{if } r = n \\ 0 & \text{if } r \neq n \end{cases} \tag{15}$$

$$\eta_{0_r} = u^{(r)}(s=0) \tag{16}$$

$$\Omega_{(\sigma_M)_{rn}} = \begin{cases} (\prod_{m=0}^{r-1}(r-m))\sigma_M^{n-r} & \text{if } n \geq r \\ 0 & \text{if } n < r \end{cases} \tag{17}$$

$$\eta_{(\sigma_M)r} = u^{(r)}(\sigma_M = 1) \tag{18}$$

Which leads to a linear, constrained quadratic minimization problem. This problem can be solved through any linear programming method in a straightforward manner namely, the elimination method or Lagrange multipliers. Introduction of a vector of Lagrange multipliers $\boldsymbol{\Lambda}$ to objective function results in the following system of linear equations:

$$\begin{bmatrix} \boldsymbol{\Phi}^T\boldsymbol{\Phi} & \boldsymbol{\Omega}^T \\ \boldsymbol{\Omega} & 0 \end{bmatrix} \begin{bmatrix} \boldsymbol{\alpha} \\ \boldsymbol{\Lambda} \end{bmatrix} = \begin{bmatrix} \boldsymbol{\Phi}^T\mathbf{u}^* \\ \boldsymbol{\eta} \end{bmatrix} \tag{19}$$

Given that all segments of arc length are non-zero, it follows that the matrix in equation (19) possesses full rank and is thus invertible. Consequently, the normalized coefficients $\boldsymbol{\alpha}$ may be readily derived via the application of the Cholesky decomposition method. To minimize the estimation of the curve parameter, a recursive adaptive step is incorporated, segmenting the polynomial into several piecewise polynomials until the mean squared error between the actual and estimated parameters falls below a specified tolerance. Should this tolerance be exceeded, the data set $\boldsymbol{u}_j^*$ is divided into two equal subsets, with a polynomial fitted to each group of points using the proposed method. This iterative process of splitting and fitting terminates when all modifier polynomials meet the specified condition or when the amount of data in a subset equals the degree of the polynomial. Upon convergence, the nonlinear relationship between $s$ and $u$ is represented by a series of piecewise ninth-order C3 continuous polynomials.

### 3.3. Tool orientation interpolation

As depicted in the flowchart presented in Figure 4, the tool orientation B-spline is not directly fitted to the raw tool orientation data, in contrast to the tool tip position. To ensure that the magnitude of the orientation vectors remains equal to unity, the tool orientation data is initially mapped into quaternion space. Given the absence of an explicit relationship between the path length parameter and the B-spline parameter of quaternions, a non-linear relationship between the arc length parameter and the quaternion parameter is established through the use of Bezier curves and non-linear optimization. Ultimately, through inverse mapping of the quaternions, the Euler angles denoting the spatial orientation of the tool during the corresponding interpolation period are derived.

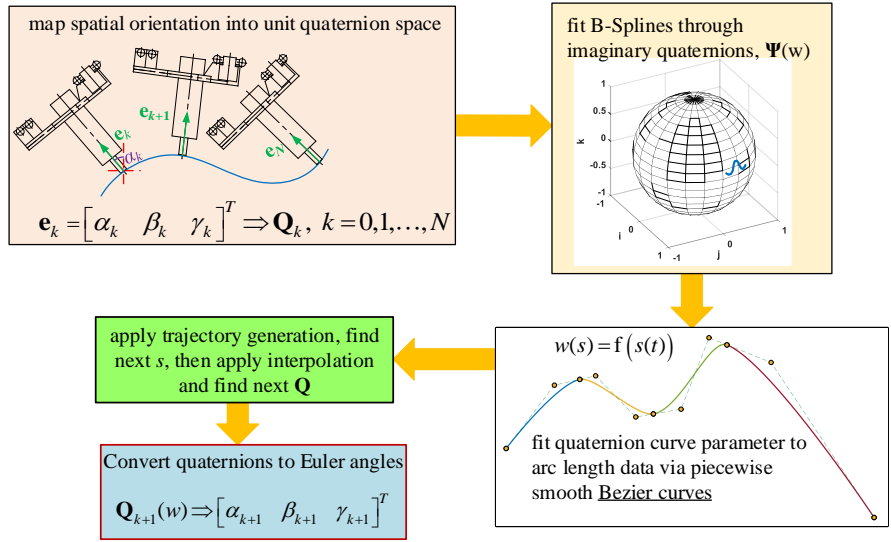

**Fig.4.** A flowchart describing the process of spatial interpolation of tool orientation. All computationally intensive steps (quaternion conversion, B-spline fitting, and nonlinear optimization for interpolation between B-Spline curve parameter and quaternions) occur offline in approximately 1.5 second for ~300 waypoints in optimized C++.

While the rotation matrix provides a convenient means of expressing the spatial orientation of a rigid body, this approach is not without its limitations. One notable issue pertains to the order of matrix multiplication; for a given set of Euler angles, different multiplication sequences can yield disparate results. Another challenge associated with the use of rotation matrices is the phenomenon of gimbal lock, which arises due to singularities in the matrix. This can result in the loss of one degree of rotational freedom for certain angular configurations of the rigid body. The aforementioned challenges, render the parametric fitting and interpolation of rotation matrices a complex task. Consequently, smooth interpolation is facilitated through the conversion of Euler angles into unit quaternions.

A unit quaternion, which is a type of hyper-complex number and satisfy the condition $\|\mathbf{Q}\|_2 = 1$, can be utilized to provide a unique description of an object's orientation. It is expressed in the following manner:

$$\mathbf{Q} = [q_0 \quad q_1 \quad q_2 \quad q_3]^T = q_0 + q_1 i + q_2 j + q_3 k \tag{20}$$

Wherein $q_r$, $r \in \{0, ..., 3\}$ are real numbers, while $i, j$, and $k$ are imaginary units and satisfy $i^2 = j^2 = k^2 = ijk = -1$. This subset of quaternions constitutes the group of unit quaternions.

In order to streamline the notation, the vector representation comprising the real component $q_0$ and the imaginary vector component $\mathbf{q}$ is employed, resulting in:

$$\mathbf{Q} = [q_0, \mathbf{q}]^T, \quad \mathbf{q} = [q_1 \quad q_2 \quad q_3] \in \mathbb{R}^3 \tag{21}$$

The multiplication of two quaternions yields a resultant quaternion, which is derived through the subsequent relationship:

$$\mathbf{Q} = \mathbf{Q}_1\mathbf{Q}_2 = [q_{10}, \mathbf{q}_1]^T[q_{20}, \mathbf{q}_2]^T = [q_{10}q_{20} - \mathbf{q}_1^T\mathbf{q}_2, \quad q_{10}\mathbf{q}_2 + q_{20}\mathbf{q}_1 + \mathbf{q}_1 \times \mathbf{q}_2]^T \tag{22}$$

Quaternion multiplication is non-commutative, meaning $\mathbf{Q}_1\mathbf{Q}_2 \neq \mathbf{Q}_2\mathbf{Q}_1$. The conjugate of a quaternion $\mathbf{Q}$ is denoted as $\mathbf{Q}^*$ and is defined as $\mathbf{Q}^* = [q_0, -\mathbf{q}]^T$.

A unit quaternion can be represented as a point on the unit hypersphere in S3 space. Quaternions can represent rotations in three-dimensional Euclidean space using their trigonometric form. Rotation of an arbitrary vector $\mathbf{p} = [x \quad y \quad z]^T \in \mathbb{R}^3$ about the unit vector $\mathbf{n} = n_1 i + n_2 j + n_3 k$ by an angle of $\varphi$ can be represented in the form of unit quaternions as follows:

$$\mathbf{p}' = \mathbf{Q}\mathbf{p}\mathbf{Q}^* \tag{23}$$

This implies that the result of two consecutive rotations can be obtained by multiplying pair of unit quaternions, which is equivalent to the overall rotation matrix. Using this reasoning, the relationships for converting between Euler angles and quaternions have been derived. Cayley's method is the most efficient approach for directly calculating the components of quaternions using the following equations:

$$q_i = \frac{1}{4}\left(\left(1 + r_{ii} - \text{tr}(\mathbf{R})\right)^2 + \left(r_{32} - (-1)^i r_{23}\right)^2 + \left(r_{13} - (-1)^i r_{31}\right)^2 + \left(r_{21} - (-1)^i r_{12}\right)^2\right)^{1/2}, \ \text{tr}(\mathbf{R}) = r_{11} + r_{22} + r_{33} \tag{24}$$

Similar to the position of the tool tip, the vector of tool orientations along the path, $\mathbf{O}_k = [\alpha_k, \beta_k, \gamma_k]^T$ must be fitted to a parametric B-spline. Similar to curve fitting for tool tip position, the purpose of curve fitting for tool orientation is to select the degree, the knot vector $\mathbf{W}$, and the control points. But, since the length of the tool remains constant, the magnitude of the orientation vector must always be equal to unity. Therefore, in contrast to fitting on the tool tip position data, the tool orientation B-spline is not directly fitted on the raw tool orientation data. Rather, to ensure that the magnitude of the orientation vectors is always equal to unity, and avoid gimbal lock, the tool orientation curve is fitted in quaternion space. According to this, by using equation (24), the spatial orientation of the tool expressed by the Euler angles are converted into their equivalent unit quaternions $\mathbf{Q}_k = [q_{ok}, \mathbf{q}_k]$.

A basic form of quaternion interpolation involves leveraging the spherical linear interpolation (SLERP) method. This approach interpolates along the geodesic of two quaternions, thus following the shortest distance between them. Specifically, the exponential-logarithmic representation of SLERP is provided as follows:

$$\mathbf{Q}(w) = \mathbf{Q}_1\exp(\log(\mathbf{Q}_1^*\mathbf{Q}_2)w), \quad w \in [0, \ 1] \tag{25}$$

Within the domain of B-Spline interpolation in three-dimensional space, a parametrization approach that involves an angle $\varphi \in \mathbb{R}$ and an axis $\mathbf{n} \in \mathbb{R}^3$ with $\|\mathbf{n}\|_2 = 1$ has proven to be beneficial. This leads to the adoption of the quaternion representation, which can be expressed as:

$$\mathbf{Q} = [\cos(\varphi/2) \quad \sin(\varphi/2)\mathbf{n}]^T = \exp\boldsymbol{\psi}, \quad \boldsymbol{\psi} = [0 \quad (\varphi/2)\mathbf{n}]^T \tag{26}$$

Wherein $\mathbf{n} = n_1 i + n_2 j + n_3 k$, denotes a unit pure quaternion. The exponential mapping procedure produces a unit quaternion that represents a rotation, which corresponds to the pure quaternion vector $\boldsymbol{\psi}$. The scaled rotation axis is subsequently denoted as $\psi = (\varphi/2)\mathbf{n}$. The inverse of equation (26), namely the logarithmic form of a quaternion, can be defined as follows:

$$\boldsymbol{\Psi} = [0 \quad \psi]^T = \log\mathbf{Q} = \log([\cos(\varphi/2) \quad \sin(\varphi/2)\mathbf{n}]^T) \tag{27}$$

The SLERP interpolation method enables rotations with a constant angular derivative, resulting in a zero second derivative. However, this technique is not ideal for generating a path that passes through more than two quaternions or connects multiple segments. The resulting path will be discontinuous. Consequently, a novel method, derived from the SLERP approach, has been introduced for generating a parametric path that considers all input quaternion data simultaneously. To ensure that the quaternions remain unit quaternions, a mapping technique is employed, which is expressed as follows:

$$\mathbf{Q}(w) = \mathbf{Q}_1\exp\left(\left[0 \quad \frac{\varphi(w)}{2}\mathbf{n}(w)\right]^T\right) = \mathbf{Q}_1\left[\cos\left(\frac{\varphi(w)}{2}\right) \quad \sin\left(\frac{\varphi(w)}{2}\right)\mathbf{n}(w)\right]^T \tag{28}$$

Equation (28) exhibits similarity to Equation (25), with the primary difference being that in Equation (38), the angle $\varphi$ and the unit rotation axis $\mathbf{n}$ are both functions of the parameter $w$. It is noteworthy that $\mathbf{Q}_1$ represents the initial orientation. By suitably adjusting $\mathbf{n}(w)$ and $\varphi(w)$, the desired path in quaternions can be obtained. The local rotational vector $\boldsymbol{\Psi}(w) = [0, \ \psi^T(w)]^T$ is formulated, taking into account that $||\mathbf{n}||_2 = 1$, and this vector is utilized for path generation. Thus, Equation (28) can be represented as follows:

$$\mathbf{Q}(w) = \mathbf{Q}_1\exp\big(\boldsymbol{\Psi}(w)\big) \tag{29}$$

The local orientation vector $\boldsymbol{\Psi}(w)$ is represented as a pure quaternion. Equation (29) allows interpolation of a B-spline curve through the imaginary component of the quaternion. The extraction of the imaginary vector $\psi(w)$ from the quaternions $\mathbf{Q}_k$, $k = 1, \dots, N+1$, is achieved using the following expression:

$$[0 \quad \psi(w_k)]^T = \log(\mathbf{Q}_1^*\mathbf{Q}_k) \tag{30}$$

In which the subsequent equations hold:

$$\begin{cases} \varphi_k(w) = 2||\boldsymbol{\Psi}_k(w)|| \\ \mathbf{n}_k(w) = \frac{2\text{Im}(\boldsymbol{\Psi}_k(w))}{\psi_k(w)} \end{cases} \tag{31}$$

Finally, the imaginary vectors obtained from equation (30), similar to the position data, can be interpolated by B-spline curves. Assuming a knot vector $\mathbf{W}$ and control points $\Theta$ consisting of pure imaginary quaternions, the B-spline curve is expressed in matrix form similar to (5) as follows:

$$\psi(w) = \mathbf{N}_p(w)\Theta \tag{32}$$

The control points can be obtained by solving a system of linear equations, using a method identical to that employed in generating position tool paths. The difference lies in the knot vector estimation, where the geodesic angular distance is utilized instead of the Euclidean distance. Upon calculating $\varphi_k(w)$ through (32), the quaternion path is attained using equations (29) through (31). In Figure 5, the B-spline curves obtained for a sample quaternion data set utilizing the two SLERP and the B-spline method introduced in this article are shown. As is evident from Figure 5, the presented approach enables high-order continuity and results in a smoother curve.

Similar to the tooltip position parameter $u$, the tool orientation parameter $w$ must also be interpolated as a function of the toolpath length $s$. However, unlike the position spline, there is no direct relationship between $s$ and the parameter $w$, as tool orientation does not affect tooltip displacement. Hence, from the available $(s_k, \bar{w}_k)$ data, a parametric curve can be fitted to represent this non-linear relationship. However, due to scarcity of data, curves with optimized internal energy functional can be employed to ensure uniformity, at least third-order continuity, and minimal oscillations. The shape-preserving property of piecewise Bezier curves makes it suitable for the task. The definition of the $n$th order Bezier curve is as follows:

$$w_k(\sigma) = \sum_{i=0}^{n} \binom{n}{i}(1-\sigma)^{n-i}\sigma^i \rho_{i,k}, \sigma = \frac{s-s_{k-1}}{s_k-s_{k-1}} \in [0,1] \tag{33}$$

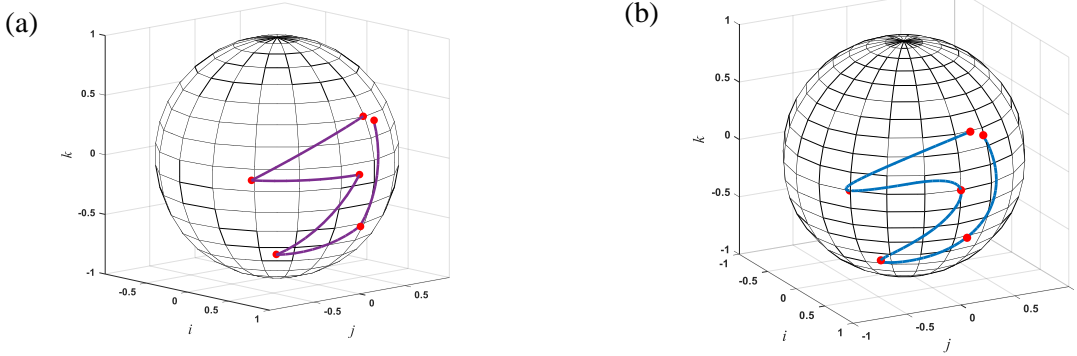

**Fig.5.** Curves obtained from the interpolation of the quaternion data using: a) SLERP, b) cubic B-spline.

In accordance with equation (33), Bezier curves are fitted to each interval$[s_{k-1}, \ s_k]$. The control coefficients $\rho_{i,k}$ are not known a priori, where the subscript $i$ denotes the numerator of the coefficients within the $k$th curve segment.

The constraints are delineated based on three distinct criteria, namely, the monotonicity of curves, the requirement that curves pass through all data points$(s_k, \overline{w}_k)$, and the need for $r$th order continuity at segment junctions. These constraints are mathematically expressed in equations (34) through (36) as follows:

$$\rho_{0,k} \le \rho_{1,k} \le \cdots \le \rho_{n-1,k} \le \rho_{n,k} \ , k = 1, \dots, N \tag{34}$$

$$[\rho_{0,k} \quad \rho_{n,k}]^T = [\overline{w}_{k-1} \quad \overline{w}_k]^T, \quad k = 1, \dots, N \tag{35}$$

$$\frac{\Delta^r \rho_{n-1,k}}{(s_k-s_{k-1})^r} = \frac{\Delta^r \rho_{1,k+1}}{(s_{k+1}-s_k)^r}, \quad \Delta^r \bar{\rho}_n = \sum_{j=0}^{r} \binom{r}{j}(-1)^{2r-j}\bar{\rho}_{n+j} \tag{36}$$

Wherein the forward difference operator, denoted by $\Delta^r$, which adheres to Pascal's triangle law.

To determine the optimal control points for piece-wise Bezier curves that interpolate the tool orientations, a nonlinear optimization problem with linear constraints may be formulated. The objective function for this problem is defined as the integral square of the $r$th derivative, subject to the constraints specified in equations (42) through (44). Specifically, for the case of $r = 3$, also known as minimized jerk, the objective function can be expressed as follows:

$$\min_{\rho_{i,k}} J_\rho = \sum_{k=1}^{N} \int_0^1 \frac{{w'''_k}^2}{(s_k-s_{k-1})^5} \, \mathrm{d}\sigma \tag{37}$$

Any nonlinear optimization method, such as interior point or active-set methods, can be employed to solve the minimization problem. Nevertheless, a straightforward time efficient method is presented in the subsequent section, given that minimum jerk trajectories are formulated in a similar manner. The solution of (37) produces the optimal control points of orientation interpolation function.

## 4. Minimum jerk time optimal Trajectory generation

Minimum-jerk Bezier splines have proven very effective as robot trajectories, since the motor commands and orientation accelerations of the mechanism are proportional to the jerk, or forth derivative of the tool path. Convex hull properties of Bezier curves can be exploited to satisfy spatial and temporal separation constraints for multi-agent trajectory generation.

This article presents a two-step optimization approach for generating minimum jerk and time optimal trajectories. The approach involves first optimizing the time allocation along each segment, followed by a feasibility-based total time optimization that takes into account drive constraints.

## 4.1 Minimum Jerk Objective function

The trajectories are generated using piece-wise Bezier curves, with arc length parameterized against time. Composite Bezier curves are formed by connecting multiple Bezier curves at their beginning and end points, while ensuring continuity of the desired degree between adjacent curves. These composite curves are defined as following:

$$
\rho(t) = \begin{cases}
\sum_{n=0}^{N} \bar{\rho}_{0,N} b_n^N \left( \frac{t-t_0}{t_1-t_0} \right) & t_0 \leq t \leq t_1 \\
\sum_{n=0}^{N} \bar{\rho}_{1,N} b_n^N \left( \frac{t-t_1}{t_2-t_1} \right) & t_1 \leq t \leq t_2 \\
\vdots \\
\sum_{n=0}^{N} \bar{\rho}_{m-1,N} b_n^N \left( \frac{t-t_{m-1}}{t_m-t_{m-1}} \right) & t_{m-1} \leq t \leq t_m
\end{cases}
\tag{38}
$$

Where in $b_n^N(\zeta) = \binom{N}{n}(1-\zeta)^{N-n}\zeta^n$ is the Bernstein basis function.

Any Bezier curve that is part of a composite curve can be defined individually with normalized time, as follows:

$$
\rho_k(\zeta) = \sum_{n=0}^{N} \bar{\rho}_n b_n^N(\zeta) \qquad \zeta = \frac{t-t_{k-1}}{\tau}, \ \tau = t_k - t_{k-1}
\tag{39}
$$

Then, integral squared norm of the $r$th derivative is introduced as the objective function of each segment in the following form:

$$
J(\tau) = \min_{\tau} \ \frac{1}{\tau^{2r-1}} \int_0^1 \left( \frac{d^r}{d\zeta^r} \rho(\zeta) \right)^2 d\zeta
\tag{40}
$$

It should be emphasized that the time allocation variable is positioned outside the integral, indicating that the control points determining the minimum jerk trajectory for a given segment are not dependent on the time allocated for that segment.

By formulating the derivative of the Bezier curve in a matrix format, the following expression is derived:

$$
\frac{d^r}{d\zeta^r} \rho(\zeta) = \left( \frac{N!}{(N-r)!} \right) \mathbf{B}_r(\zeta)^T \mathbf{D}_r \bar{\mathbf{P}},
\tag{41}
$$

Wherein the Bernstein matrix, the matrix of control points and the matrix containing the forward difference operators, are respectively defined as follows:

$$
\mathbf{B}_r(\zeta) = [b_0^{N-1}(\zeta) \quad b_1^{N-1}(\zeta) \quad \cdots \quad b_{N-1}^{N-1}(\zeta)]^T
\tag{42}
$$

$$
\bar{\mathbf{P}} = [\bar{\rho}_0 \quad \bar{\rho}_1 \quad \cdots \quad \bar{\rho}_N]^T
\tag{43}
$$

Furthermore, as indicated in equation (41), the matrix $\mathbf{D}_r$ is a ($N$-$r$) by $N$ matrix consisting of a forward difference operator, with non-zero elements.

The matrix relationships derived and the independence of the control points and the forward difference operator from time changes allow for the removal of these matrices from within the integral. As a result, the objective function expressed in equation (40) can be reformulated in matrix form as follows:

$$J(\tau) = \min_{\tau} \frac{1}{\tau^{2r-1}} \left(\frac{N!}{(N-r)!}\right)^2 \overline{\mathbf{P}}^T \mathbf{D}_r{}^T \left(\int_0^1 (\mathbf{B}_r(\zeta)\mathbf{B}_r(\zeta)^T) d\zeta\right) \mathbf{D}_r \overline{\mathbf{P}} \tag{44}$$

As can be seen, the expression inside the integral represents the matrix symmetric square of Bernstein polynomials. This integral can be evaluated element-wise. The integration of $F_r$ on an element-wise basis requires the utilization of the Beta and Gamma function integral theorems. Upon the completion of the integration, the objective function (44) can be represented solely as the following matrix product:

$$J(\tau) = \min_{\tau} (\prod_{k=0}^{r-1} (N-k))^2 \tau^{1-2r} \overline{\mathbf{P}}^T \mathbf{D}_r{}^T \mathbf{F}_r \mathbf{D}_r \overline{\mathbf{P}}. \tag{45}$$

The objective function of m Bezier curve segments can be evaluated by concatenating them through the Kronecker tensor multiplication rule as:

$$\overbrace{J(\tau_0, \dots, \tau_{m-1})}^{\mathbf{J}} =$$

$$\min_{(\tau_0, \dots, \tau_{m-1})} \overbrace{[\overline{\rho}_0 \quad \cdots \quad \overline{\rho}_{m-1}]}^{\overline{\mathbf{P}}} \operatorname{diag}(\overbrace{\left[\frac{1}{\tau_0^{2r-1}} \mathbf{D}_r{}^T \mathbf{F}_r \mathbf{D}_r \quad \cdots \quad \frac{1}{\tau_{m-1}^{2r-1}} \mathbf{D}_r{}^T \mathbf{F}_r \mathbf{D}_r\right]}^{\mathbf{Q}_r}) \overbrace{[\overline{\rho}_0 \quad \cdots \quad \overline{\rho}_{m-1}]}^{\overline{\mathbf{P}}} \tag{46}$$

The cost function for the entire trajectory can be expressed using a quadratic programming approach as follows:

$$\mathbf{J} = \min_{\tau} \overline{\mathbf{P}}^T \mathbf{Q}_r \overline{\mathbf{P}} \tag{47}$$

Wherein the associated segment-wise time allocation variable $\boldsymbol{\tau} = [\tau_0, \ \tau_1, \dots, \tau_{m-1}]$ is incorporated in the expression.

## 4.2. Constraints

The optimization of the piecewise Bezier curves requires the satisfaction of certain constraints, including the fixed junction points, fixed derivatives at the initial and final location based on the required initial and final states, and matched derivatives at all intermediate junction points up to the $r$th derivative to ensure the $C^r$ continuity of the curve. These constraints can be expressed in matrix form as follows:

$$A = [A_0 \quad A_1]^T = [\mathbf{B}_0^T(0)\mathbf{D}_0 \quad \cdots \quad \mathbf{B}_r^T(0)\mathbf{D}_r \quad \mathbf{B}_0^T(1)\mathbf{D}_0 \quad \cdots \quad \mathbf{B}_r^T(1)\mathbf{D}_r]^T, \tag{48}$$

In the given context, where A represents the Bernstein component of each derivative at the fixed control points, and

$$d = [d_0 \quad d_1]^T = \left[\frac{\partial^0}{\partial t^0}\rho(0) \quad \cdots \quad \frac{\partial^r}{\partial t^r}\rho(0) \quad \frac{\partial^0}{\partial t^0}\rho(1) \quad \cdots \quad \frac{\partial^r}{\partial t^r}\rho(1)\right]^T, \tag{49}$$

The vectors in (48) and (49) comprises the values corresponding to the fixed or matched derivatives at the fixed control points. It is important to note that the aforementioned constraints apply to each individual Bezier segment. To represent the constraints for the entire trajectory, the entries are populated in a block-diagonal manner as follows:

$$\overbrace{\begin{bmatrix} A_0 & & & \\ A_1 & & & \\ A_1 & -A_0 & & \\ & A_1 & & \\ & \vdots & \ddots & \\ & & & A_1 \end{bmatrix}}^{\mathbf{A}} \overbrace{\begin{bmatrix} \overline{\rho}_0 \\ \overline{\rho}_1 \\ \vdots \\ \vdots \\ \overline{\rho}_{m-1} \end{bmatrix}}^{\overline{\mathbf{P}}} = \overbrace{\begin{bmatrix} b_{0,0} \\ b_{0,1} \\ 0 \\ b_{1,1} \\ \vdots \\ b_{m-1,1} \end{bmatrix}}^{\mathbf{b}}, \tag{50}$$

### 4.3. Unconstrained QP Solution

To prevent the occurrence of singular or poorly conditioned matrices during the nonlinear optimization process, which is commonly challenging due to the large sparse matrices defining the constraint equations, a strategy can be employed. This strategy involves incorporating the constraints directly into the cost function using matrix inversion. By adopting this approach, the optimization process becomes more efficient, allowing for the direct extraction of the optimal control points based on the given time allocation. Hence, the cost function of the unconstrained formulation can be represented as follows:

$$\mathbf{J} = \min_{\boldsymbol{\tau}} \mathbf{d}^T \mathbf{A}^{-T} \mathbf{Q}_r \mathbf{A}^T \mathbf{d} \tag{51}$$

The newly formulated quadratic cost function now involves the endpoint derivatives of the segments as decision variables. These variables are re-arranged such that the fixed or specified derivatives are grouped together (denoted as $\mathbf{d}_k$), while the free or unspecified derivatives are grouped together (denoted as $\mathbf{d}_u$). To achieve this re-ordering, a sparse permutation matrix ($\mathbf{M}$) is constructed. Thus, the revised expression is as follows:

$$\mathbf{J} = \min_{\boldsymbol{\tau}} \begin{bmatrix} \mathbf{d}_k \\ \mathbf{d}_u \end{bmatrix}^T \overbrace{\mathbf{M}\mathbf{A}^{-T}\mathbf{Q}_r\mathbf{A}^{-1}\mathbf{M}^T}^{\mathbf{R}} \begin{bmatrix} \mathbf{d}_k \\ \mathbf{d}_u \end{bmatrix} = \min_{\boldsymbol{\tau}} \begin{bmatrix} \mathbf{d}_k \\ \mathbf{d}_u \end{bmatrix}^T \begin{bmatrix} \mathbf{R}_{kk} & \mathbf{R}_{ku} \\ \mathbf{R}_{uk} & \mathbf{R}_{uu} \end{bmatrix} \begin{bmatrix} \mathbf{d}_k \\ \mathbf{d}_u \end{bmatrix} \tag{52}$$

The augmented cost matrix is subsequently partitioned based on the indices of the fixed and free derivatives.

By differentiating $\mathbf{J}$ and setting it to zero, the vector of optimal values for the free derivatives can be obtained in terms of the fixed or specified derivatives and the cost matrix. The equation representing this relationship is as follows:

$$\mathbf{d}_u^* = -(\mathbf{R}_{uu}^T + \mathbf{R}_{uu})^{-1}(\mathbf{R}_{ku}^T + \mathbf{R}_{uk})\mathbf{d}_k \tag{53}$$

Subsequently, the optimal control points for a specific time allocation can be computed using the following equation:

$$\overline{\mathbf{p}}^* = \mathbf{A}^{-1}\mathbf{M}^T \begin{bmatrix} \mathbf{d}_k \\ \mathbf{d}_u^* \end{bmatrix} \tag{54}$$

The configuration of the control points remains unaltered irrespective of the scaling in the time allocation since the time variable does not contribute to the integral cost for an independent Bezier segment. To determine the optimal arrangement of control points, the total sum of allocated times is constrained to unity, represented as $\tau_n = (\tau_0, \ \tau_1, \dots, \tau_{m-2}, \ 1 - \sum_{k=0}^{m-2} \tau_k)$ ensuring that every time allocation value is consistently positive. This procedure is repeated in the actuator space

### 4.4. Optimal Time Trajectory

Since the optimized trajectory is parametrized within the range of $t$ [0, 1], regardless of the scale of the solution, the resulting trajectories may not always be feasible. To address this, a secondary optimization is conducted to determine the optimal total flight time that adheres to the actuator constraints imposed by the mechanism. The cost function $\mathbf{J}_T$, representing the total flight time, can be formulated as a constrained optimization problem, subjected to tangential and axial kinematic constraints as shown below:

$$\mathbf{J}_T = \min_k \quad \mathbf{J}(k\tau),$$
$$\text{s.t} \quad \begin{cases} |\ddot{s}| < a_{\max} \ , & |\dddot{s}| < j_{\max} \\ |\ddot{d}| < a_{d_{\max}} & |\dddot{d}| < j_{d_{\max}} \end{cases} \tag{55}$$

The optimization problem can be solved using nonlinear optimization methods such as interior point or active set methods. In this particular study, the open source NLopt C++ library is utilized to solve equation (55).

### 4.5. Actuator space layer interpolation

In order to determine the spatial position of the tool during each interpolation period and the corresponding length of the traversed path, a series of calculations are performed. Initially, the B-spline curve parameter is computed, which allows for the determination of the B-spline curve position. If there are changes in the spatial orientation of the tool, the imaginary part of the unit quaternions is calculated, and subsequently, the quaternion path is derived. Finally, the obtained quaternions are transformed back to Euler angles, providing the position and spatial orientation of the tool relative to the vector **P**, at time *t*. The inverse kinematics is then employed to obtain the displacement vector for the actuators.

Subsequently, the minimum snap/jerk trajectory generation method, previously discussed for the task space, is further extended to the actuator space. This entails determining the coefficients of the polynomials that describe the motion of the actuators. Once the coefficients are obtained, they are transmitted to the microprocessor during each interpolation interval, facilitating the generation of the requisite pulses for the accurate movement of the motors.

The proposed trajectory-generation framework consists of two main computational stages: (1) offline preprocessing, which includes numerical integration for position interpolation, nonlinear optimization for orientation interpolation, and time-optimal trajectory generation; and (2) real-time execution, where precomputed coefficients are used for efficient interpolation.

The offline preprocessing stage, which is executed once per trajectory, requires a total computation time of approximately 5 seconds for a trajectory consisting of 300 waypoints, broken down as follows: position interpolation via numerical integration (~3.3 sec), orientation interpolation via nonlinear optimization (~0.4 sec), and minimum jerk time optimal trajectory generation (~0.4 sec).

The real-time execution stage is computationally efficient due to precomputed coefficients. The total time per real-time interpolation step is approximately 65 µs, making the proposed method highly suitable for real-time industrial robotic applications.

### 5. Results and Discussion

Figure 6 presents the motion unit's configuration, which comprises four distinct components: a graphical user interface (GUI), microprocessors, motors, and linear encoders. Within the computer system, intensive calculations such as the parametric interpolation of the tool path and motion planning in Cartesian space are executed using object-oriented C++ programming. The system ultimately produces a set of polynomial coefficients that define the motion equations in actuator space. In addition, a Qt-based user interface has been developed to enable seamless hardware interaction, allowing users to perform operations such as referencing, issuing movement commands, and executing stop commands with ease.

A 32-bit ARM-based microprocessor (Arduino Due) is responsible for generating movement pulses and transmitting them to the motor drives. Furthermore, four AVR-based Arduino Mega 2560 microprocessors are deployed to receive signals from the encoders, process the data, and relay the results to the GUI. Once the digital pulses for motion and direction are received, the drivers amplify the voltage and perform the necessary processing to ensure that the motion signals are accurately transmitted to the motors, thereby enabling precise rotational movement.

The optimized C code running on the Arduino Due has achieved an interpolation loop time of 55 microseconds, approaching real-time performance. All tool path generation and trajectory optimization are carried out on a PC (ASUS N43sn laptop, 2nd generation 2.2 GHz CPU). The piecewise trajectory polynomial coefficients for all joint space segments are stored in a lookup table and transmitted to the microcontroller via UDP communication; the only online operation is the calculation of the next step at each timer interrupt to drive the motor to its desired pose at the corresponding time.

The system continuously monitors incoming data from network communications and stores the received data in buffers. It then generates and transmits precise digital pulses to the motor driver. To achieve this, an internal timer within the central processor, operating at a frequency of 84 MHz, is employed. This timer triggers system interrupts for brief durations during which all computations related to pulse generation and transmission are executed within a dedicated function. Rigorous benchmarking determined that the total time required to calculate the displacement of four motors using eighth-degree polynomials is at least 55 microseconds. To accommodate additional processor tasks between interrupts, a slight temporal buffer of a few extra microseconds is incorporated. Consequently, the interrupt timer is set to 65 microseconds, yielding 0.4 MHz in algorithm frequency.

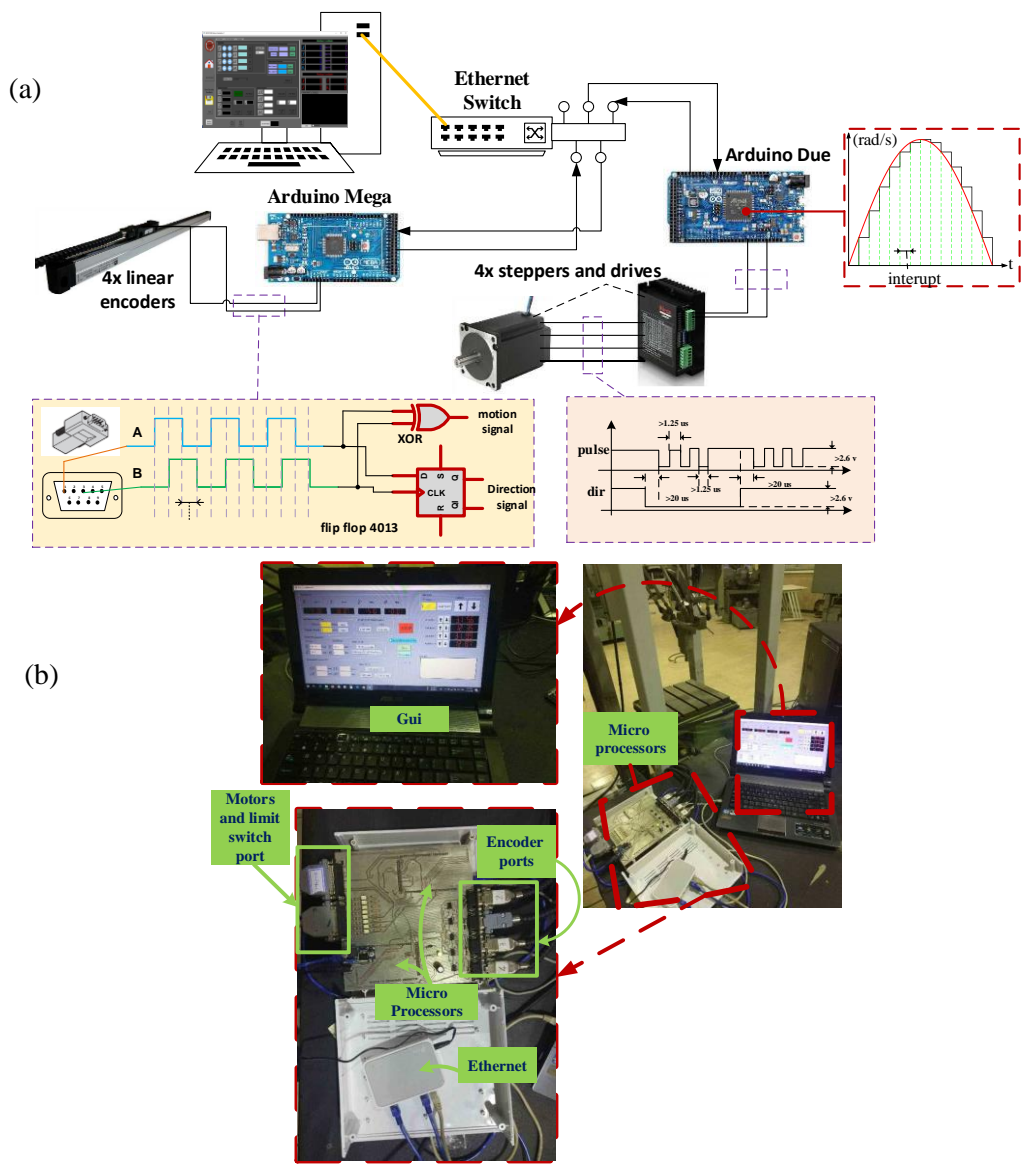

**Fig.6.** Motion generation unit: (a) schematic overview (b) real-world implementation.

The initial test involved a fan-shaped tool path comprising 89 data points, which were subsequently scaled to an approximate size of 125 (mm) by 125 (mm). As depicted in part (a) of Figure 7, a 5th degree B-spline curve was fitted to these data points using the methodology outlined in the preceding section. The resulting tool path exhibited a total length of 568.268 (mm). Examination of the curvature graph plotted against the curve parameter, as presented in part (b) of Figure 7, revealed the presence of four sharp corners within the tool path. Subsequently, a comparative analysis was conducted to assess the performance of the interpolation and trajectory generation method proposed in this research. In this study, two distinct performance indices related to trajectory smoothness and accuracy are explicitly defined and analyzed. First, feed-rate fluctuation explicitly refers to sudden or rapid variations of instantaneous feed-rate over short periods. Such fluctuations are quantitatively characterized by peak jerk values, Root Mean Square (RMS), and standard deviation of higher-order derivatives (such as jerk). Large fluctuations represent abrupt and frequent feed-rate changes, which cause mechanical vibrations, increased actuator stress, and reduced motion precision. Second, the term feed-rate deviation (tracking error) explicitly indicates the instantaneous absolute difference between actual feed-rate values measured experimentally through encoder data and direct kinematics, and analytically calculated ideal feed-rates. This second metric primarily reflects how accurately the trajectory execution adheres to the ideal planned trajectory.

To facilitate this analysis, motion planning was conducted employing the minimum jerk approach and a maximum speed of 20 (mm/s), with a maximum acceleration of 300 (mm/s²).

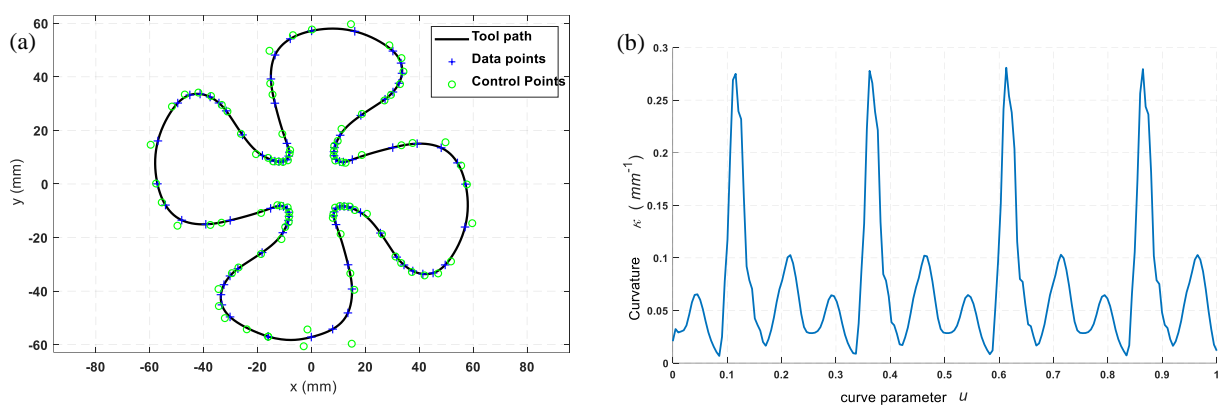

**Fig.7.** (a) Fan-shaped B-spline path, (b) curvature variation against curve parameter.

In order to quantitatively demonstrate the effectiveness of the proposed optimal trajectory generation method, Figure 8 presents a comparison between the optimized time-optimal minimum-jerk trajectory and a conventional piecewise polynomial trajectory of identical polynomial degree (7th degree). The comparisons are shown through three key motion metrics: feed-rate, acceleration, and jerk. As observed from Figure 8(a) and further supported by statistical analysis, the optimized trajectory significantly reduces feed-rate fluctuations, with peak feed-rate decreasing by approximately 73.4% (from 70.70 mm/s to 18.84 mm/s) compared to the non-optimal trajectory. Furthermore, the optimized method yields a 79.7% reduction in feed-rate standard deviation (from 13.55 mm/s to 2.76 mm/s), indicating a consistently smoother speed profile throughout the trajectory.

Acceleration profiles, shown in Figure 8(b), illustrate even greater improvements. The optimized trajectory achieves an 88.6% reduction in peak acceleration (from 260.16 mm/s² down to 29.65 mm/s²) and an 89.8% reduction in both RMS and standard deviation values (from 51.71 mm/s² to 5.30 mm/s²). Such considerable reductions directly translate to diminished mechanical stress, reduced vibrations, and increased trajectory tracking precision. Figure 8(c) emphasizes jerk, a critical indicator of motion smoothness, illustrating the most substantial improvement. Peak jerk is reduced by 95.7% (from 1793.82 mm/s³ to 77.84 mm/s³) in the optimized method. Additionally, jerk RMS and standard deviation both show reductions of approximately 95.0% (from 311.04 mm/s³ to 15.61 mm/s³), confirming drastically smoother acceleration transitions.

Collectively, these statistical outcomes clearly substantiate the practical superiority of the optimized trajectory generation approach. By achieving substantially lower peak values, RMS magnitudes, and variations in jerk, acceleration, and feed-rate, the proposed method effectively balances fast execution with minimal dynamic disturbances. This leads to improved mechanical stability, enhanced surface finish quality, decreased machine wear, and potentially reduced energy consumption in manufacturing and robotic machining processes.

Regarding computational efficiency, the proposed optimal trajectory-generation approach, demonstrates highly favorable computational complexity. Traditional dense polynomial trajectory optimization algorithms exhibit cubic complexity ($O(n^3)$). However, by adopting a sparse unconstrained quadratic-programming formulation with optimized sparse linear algebra routines, the proposed method significantly reduces practical complexity to approximately quadratic ($O(n^3)$). For example, benchmarks clearly illustrate solution times of approximately 0.34 ms for a 3-segment trajectory using C++/Eigen dense solvers. Extending these benchmarks explicitly to a larger scenario involving, for instance, 300 waypoints (~100 polynomial segments), would naively lead to approximately a 100-fold increase in complexity. This would yield an estimated computational time of around 340 ms, maintaining excellent scalability and feasibility for robotic applications. Furthermore, due to the offline calculation and storage of polynomial coefficients, the real-time computational overhead per interpolation step is minimal (on the order of 65 μs per step on the micro-controller side), explicitly reinforcing the real-time suitability of this method.

To evaluate the effectiveness of the proposed interpolation method presented in this article, an assessment was conducted by comparing the obtained results with the direct kinematic output from encoders. Table.1 lists the feed-rate, acceleration, and jerk values derived from the analytical relations, juxtaposed with the corresponding experimental values obtained from application of direct kinematics to encoder data. In the respective sections of this table, the outcomes obtained from natural interpolation methods, Taylor expansion of first-order and second-order, as well as modifier polynomials with varying tolerances of least square error (specifically, 1e-8, 1e-10, and 1e-12), have been listed. This comparison provides insights into the performance of the proposed interpolation method in relation to these alternative approaches.

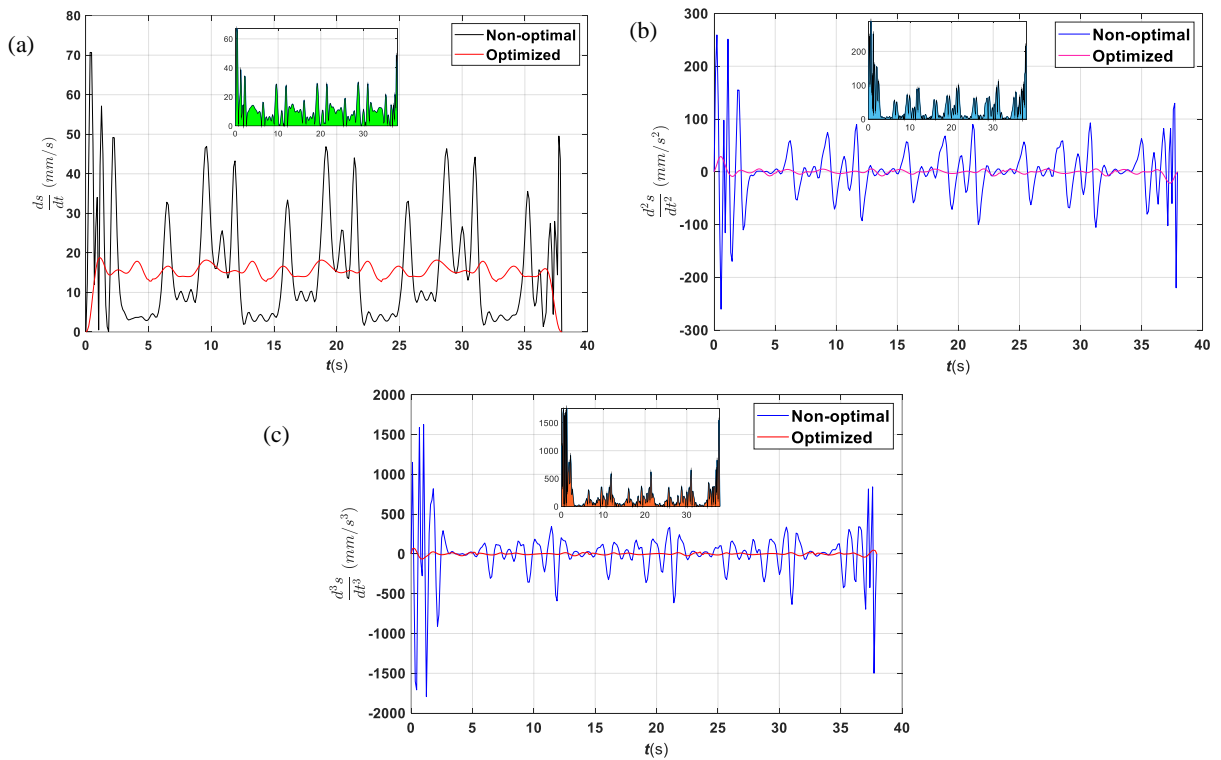

**Fig.8.** comparison of presented optimum and general piece-wise non-optimal trajectory for 7th degree

Bezier Curves: **(a)** Feed-rate, **(b)** Acceleration, **(c)** Jerk.

Based on Table 1, it is evident that the natural interpolation method exhibits significant feed-rate fluctuations. Consequently, the maximum feed-rate during path travel exceeds the predetermined limit, reaching approximately 30 (mm/s). Moreover, the maximum deviation from the ideal feed-rate is approximately 12 (mm/s), with an average fluctuation of 4.6 (mm/s) during motion. This behavior arises due to the assumption of a linear relationship between the curve parameter and a fraction of the path length in the natural interpolation method, which fails to account for the non-linear relationship between the path length and the parameter of the B-spline curve. As discussed in Section 3, achieving a perfectly linear relationship between displacement and curve parameter is only possible for straight lines without any curvature. Consequently, part (b) of Figure 7 demonstrates that the presence of curvature and significant changes in curvature renders the natural interpolation method inaccurate for free-form curves. Furthermore, the actual acceleration and jerk using the natural interpolation method deviates significantly from the ideal state. The observed large sudden changes in acceleration and jerk can be attributed to the assumption of a linear relationship between the curve parameter and the path length in the natural interpolation method, which fails to capture the higher-order continuity of the curve parameter with respect to the path length.

**Table.1.** Comparative analysis of interpolation methods based on motion indices. The table summarizes the absolute deviations of feed-rate, acceleration, and jerk from their ideal values, as well as the computation time for each interpolation method.

| Interpolation Method | Feed-rate deviation (mm/sec) | | Acceleration deviation (mm/sec$^2$) | | Jerk deviation (mm/sec$^3$) | |
|---|---|---|---|---|---|---|
| | max | mean | max | mean | max | mean |
| Natural | 11.852 | 4.467 | 38.047 | 8.808 | 115.554 | 30.930 |
| 1st Taylor's | 2.755 | 0.568 | 14.955 | 2.065 | 91.953 | 9.722 |
| 2nd Taylor's | 1.648 | 0.137 | 4.490 | 0.630 | 41.477 | 4.230 |
| Modifier polys $\varepsilon_{MSE}$=1e-8 | 0.809 | 0.064 | 3.300 | 0.387 | 37.582 | 2.833 |
| Modifier polys $\varepsilon_{MSE}$=1e-10 | 0.676 | 0.047 | 3.277 | 0.300 | 35.134 | 2.514 |
| Modifier polys $\varepsilon_{MSE}$=1e-12 | 0.642 | 0.037 | 2.932 | 0.236 | 20.791 | 1.867 |

Regarding the first-order Taylor expansion interpolation method, referring to the results listed in Table 1 show a notable improvement compared to the natural interpolation method in terms of reducing the deviation of feed-rate from the ideal values. Fluctuations in feed-rate are observed primarily in four regions characterized by sharp corners, attributed to the high curvature of the curve within these zones. The impact of curvature on the fluctuations is expected, as the curve's slope varies significantly in high-curvature points, resulting in inaccurate approximation of the curve parameter with respect to the path length. In the first-order Taylor expansion method, the curve parameter is assumed to have a linear relationship with the first-order derivative of the path length function, disregarding higher-order derivatives. Consequently, round-off errors, as well as substantial errors in approximating the curve parameter in regions with large second derivatives (i.e., higher curvature), lead to feed-rate fluctuations. An additional drawback of the first-order Taylor expansion method is its time-consuming nature, as it requires analytical calculations of the first-order derivative of the B-spline curve in each interpolation cycle. Results from the second-order Taylor expansion method, listed in Table 1, exhibits significant improvement in the proximity of the actual feed-rate to the ideal value. However, the lack of high-order continuity boundary conditions between the segments and cumulative rounding errors leads to feed-rate fluctuations. Despite the improved smoothness and proximity to the ideal state compared to the first-order expansion method, the computation time for the second-order Taylor expansion method is high due to the requirement of calculating the first and second-order derivatives of the curve during interpolation.

The modifier polynomial method, consistently yields superior outcomes compared to other interpolation methods. As expounded upon in Section 3, this technique leverages the analytical relationship between the path length and the integral of the first-order derivative of the underlying B-spline curve. It formulates a linear optimization problem aimed at minimizing the least squared error between the estimated curve parameters and their ideal values, while incorporating continuity boundary conditions of at least third order. The resulting algorithm generates coefficients for multiple ninth-degree polynomials, each approximating the curve parameter within a specific interval during each interpolation period, based on the known path length. The number of polynomials utilized depends on the specified tolerance for the least squares error. As the approximate curve parameter aligns more closely with the ideal value, variations in the path length, calculated via motion relationships derived from minimized jerk, increasingly match the patterns of motion curves, resulting in minimal feed-rate oscillations along the path. As the tolerance for the least squares error tightens the actual motion curves follow the desired trend more closely. The modifier polynomial interpolation method, employing a tolerance of $\varepsilon_{MSE} = 1e - 8$ using 1655 path length data obtained from Simpson's integral, 31 piecewise polynomials are obtained. even with a tolerance of 1e-8, this interpolation method exhibits smoother motion than the second-order Taylor expansion method. Further reduction in the least squares error tolerance leads to enhanced quality and smoothness of the motion curves. The corresponding number of polynomials for these tolerances is 33 and 63, respectively. Another advantageous aspect of the modifier polynomial method, compared to Taylor expansion approaches, lies in computation time. Since the modifier polynomial coefficients are preprocessed and stored in tables, the interpolation process experiences significant reduction in calculation time, estimated to be around 65 us for Atmel SAM3X8E ARM Cortex-M3.

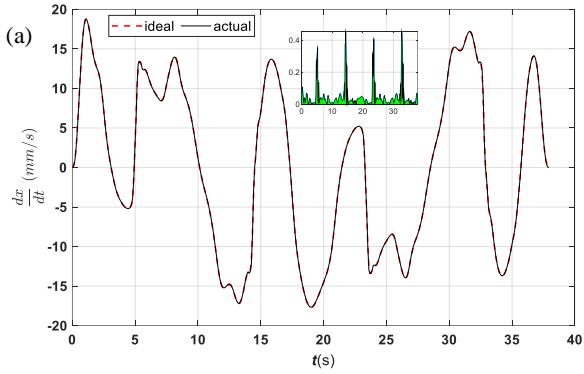
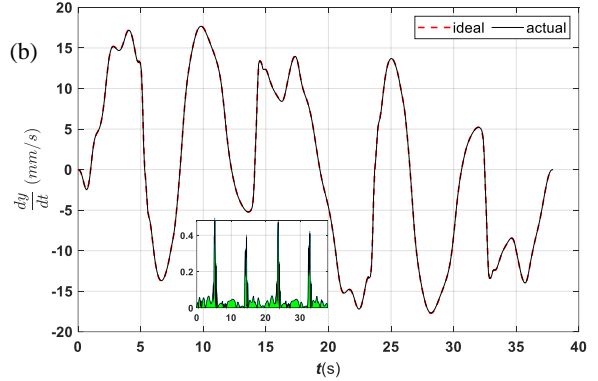
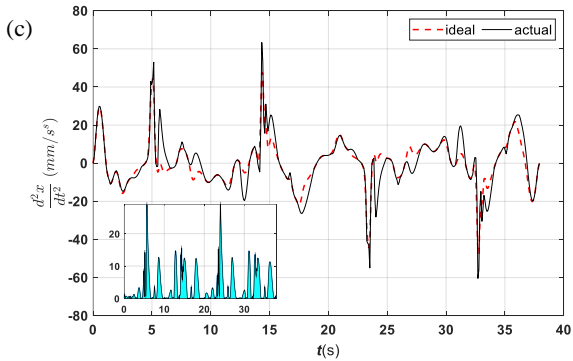
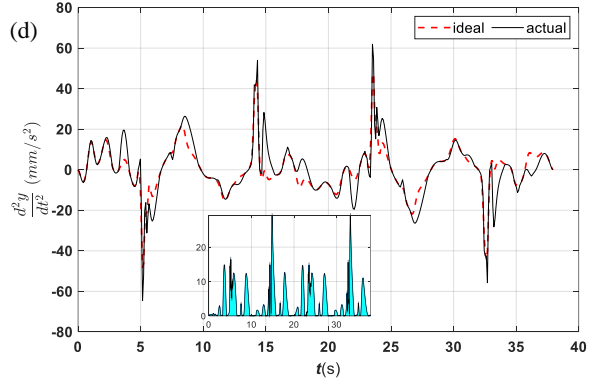

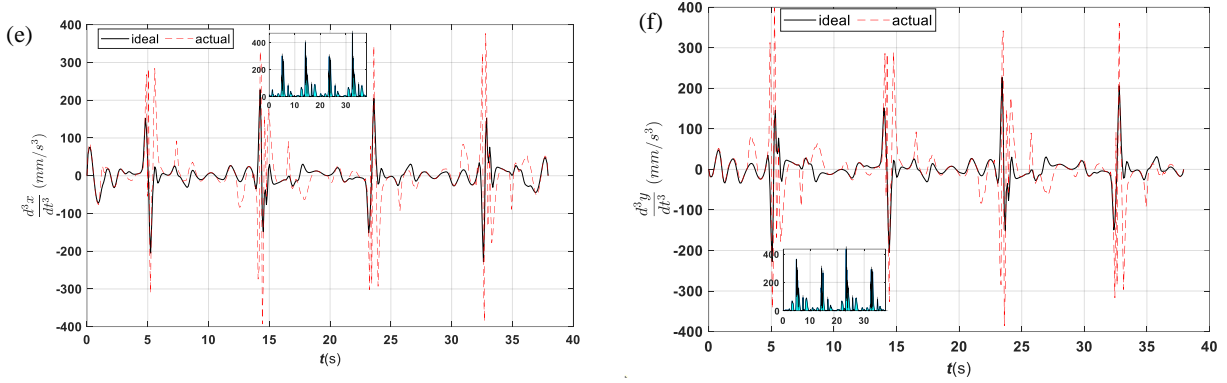

**Fig.9.** comparison of ideal and actual motion profiles for end-effector in cartesian space: **(a)** x axis feed-rate, **(b)** y axis feed-rate, **(c)** x axis acceleration, **(d)** y axis acceleration, **(e)** x axis jerk, **(f)** y axis jerk.

Figure 9 presents a comparison between the desired (planned) trajectories and actual (measured) trajectories obtained via direct kinematic calculations using encoder data. It can be clearly observed that the actual trajectories closely follow the desired profiles, confirming the high accuracy and reliability of the proposed interpolation method. However, subtle deviations occur, especially in regions associated with sharp directional changes or significant curvature variations. These slight deviations between planned and actual paths arise primarily from inherent numerical approximations and limited dynamic response of the robotic system in real-time execution.

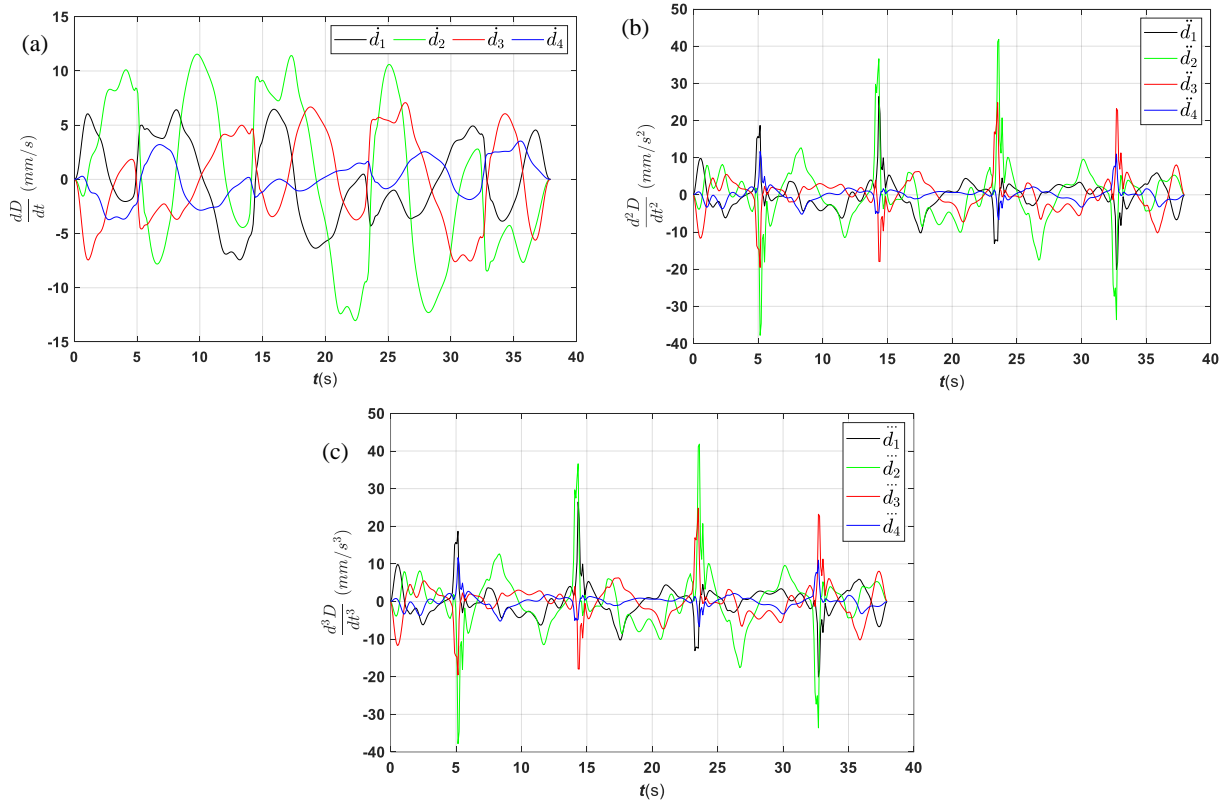

**Fig.10.** Resultant actual motion profiles for joint space via the presented optimal time minimum jerk trajectory generation: **(a)** velocity, **(b)** acceleration, **(c)** jerk.

To further investigate these deviations, Figure 10 explicitly depicts the corresponding joint-space profiles, including velocity, acceleration, and jerk, for each joint. Although the Cartesian-space trajectories appear generally smooth and close to the desired trajectory, the joint-space results illustrate clear dynamic complexities. Specifically, joints frequently change directions, driven by the nonlinear mapping of Cartesian-space paths to joint angles through inverse kinematics. As the robot end-effector moves smoothly along the planned Cartesian path, individual joints must continually alter their velocities and directions to precisely follow toll path, inherently resulting in notable acceleration and jerk spikes. These joint-level dynamic variations are primarily a consequence of nonlinear mapping from Cartesian to joint space, as joint axes constantly reorient and accelerate to accommodate smooth end-effector movements.

The presence of these acceleration and jerk spikes at the joint level clearly underscores the necessity of explicitly constraining joint-level dynamics during trajectory planning, particularly for parallel robotic mechanisms. Nonetheless, these peaks remain within acceptable actuator and mechanical limits, validating the effectiveness of the proposed trajectory-generation method in managing axis-level dynamics explicitly.

Moreover, the machined part is illustrated in part (a) of Figure 11. Data points from the machined curve were extracted using a profile projector device, as depicted in part (b) of Figure 11. By selecting a reference point, 200 points were extracted from both the outer and inner edges of the machined curve. To reconstruct the path, a B-spline curve was fitted to these data points by taking the median, following the procedure described in Section 3. The results, as shown in parts (a) and (b) of Figure 12, indicate that despite compensating for the movement error of the sliders, errors of approximately hundredths of a millimeter were still observed in the x and y directions. These discrepancies may be attributed to various factors, including errors caused by the expansion coefficient of rails, joints, arms, and clearance between the arms.

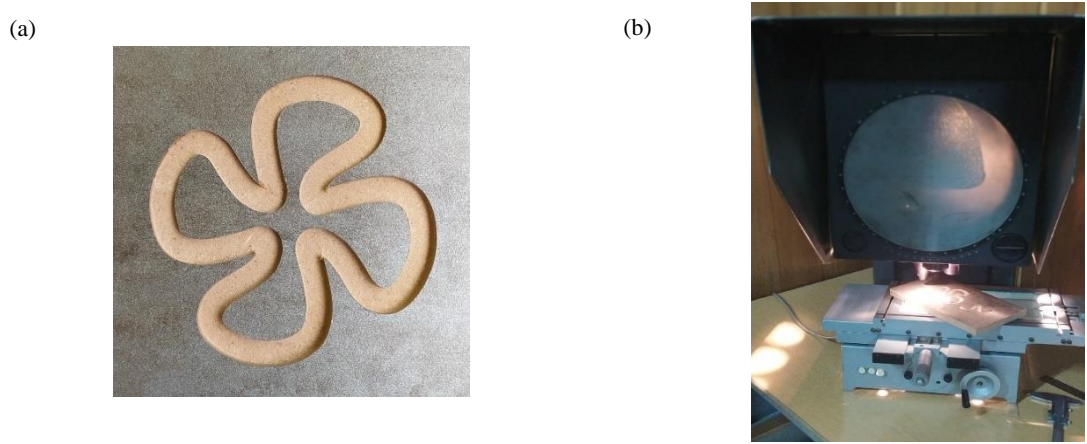

(a)                                        (b)

**Fig.11. (a)** A visual representation of the machined freeform fan-shaped tool path, and **(b)** An illustration of the curve profile measurement setup using the projector profile device.

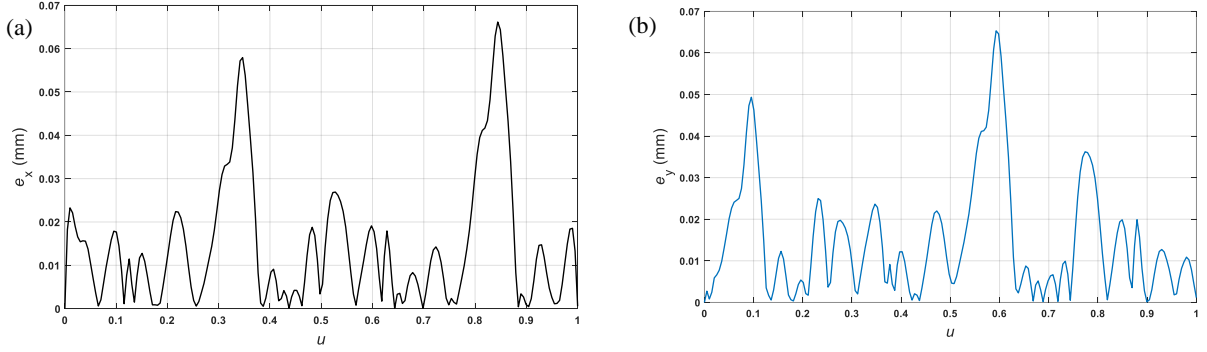

**Fig.12.** The error arising from the disparity between experimental and simulation data for the fan-shaped machined part: **(a)** along the x-axis, and **(b)** along the y-axis.

To validate the tool path orientation, a curved path was derived from 18 data points representing a spherical section in the y-z plane. Following the procedure detailed in Section 3, a fifth-degree B-spline curve was fitted to these data points. The resulting curve, illustrated in Figure 13, defines the tool path. Tool orientation was established by ensuring that the tool axis remained perpendicular to the path. This perpendicular vector was computed as the second derivative of the B-spline curve with respect to the geometric parameter, following the derivation presented in Equation 8. Euler angles were determined by calculating the angle between the tool's initial orientation vector and vectors perpendicular to the fitted curve. Using the improved Simpson integration method, the curve was discretized into 103 segments, resulting in a total path length of 63.511 mm. Position interpolation for the tool tip was executed using 14 piecewise-modified ninth-degree polynomials, with the interpolation tolerance set to $\varepsilon_{MSE} = 1e - 12$, ensuring high accuracy in the approximation of the curve parameter along the path.

The tool's spatial orientation was represented by converting Euler angles into unit quaternions within a four-dimensional quaternion space. This conversion employed a fifth-degree B-spline fitting technique, as described in Subsection 3.3. Subsequently, spatial orientation was parameterized through the optimization of piecewise Bezier curves relative to arc length. To achieve smooth temporal parameterization of the path length, the trajectory generation approach described in Section 4 was utilized. Given an offline interpolation interval of 10 ms, actuator displacements were calculated via inverse kinematics. Each displacement segment underwent trajectory generation within actuator space, following the same procedure.

Upon trajectory calculation, segment data was transmitted to the microcontroller, which executed motion generation at a 65 μs interpolation cycle. After motion completion, linear and angular positions, velocities, accelerations, and jerk values of the tool were computed using direct kinematics, with encoder feedback ensuring accuracy.

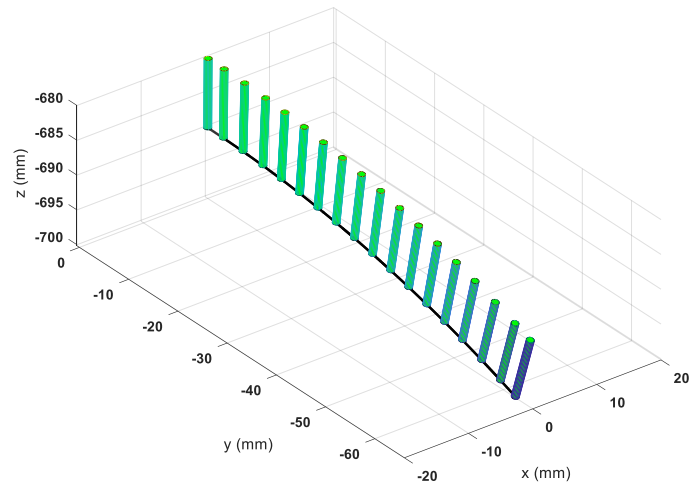

**Fig.13.** B-spline toolpath reconstructed from point cloud, where the tool axis maintains a consistent perpendicular orientation to the path.

Figure 14 illustrates the velocity, acceleration, and jerk resulting from the motion along the tool path. The curves for velocity, acceleration, and jerk exhibit minimal deviations from their ideal counterparts due to the low curvature and small changes in curvature along the path. The maximum feed-rate deviation, as shown in part (a) of Figure 19, is 0.009 (mm/s) with an average of 0.001 (mm/s) along the path. This demonstrates the high performance of the interpolation method with modified polynomials in curved paths with low curvature. The trajectory generation with minimum jerk ensures minimal acceleration changes while satisfying continuity constraints. Consequently, a nearly uniform acceleration is achieved after the initial acceleration phase, resulting in a smooth and continuous motion. Part (b) of Figure 19 shows the maximum deviation from the ideal acceleration, which is 0.1904 (mm/s²) with an average deviation of 0.0165 (mm/s²) along the route. The acceleration curve exhibits smooth variations without sharp fluctuations, reflecting the influence of the minimum jerk trajectory profile defined by third-order continuity. Part (c) of Figure 19 displays the jerk curve obtained from the encoder feedback and direct kinematics compared to the ideal state. The maximum jerk deviation is reported as 7.123 (mm/s³) with an average deviation of 0.4 (mm/s³) along the route. The observed variations in jerk are predominantly present in areas where the curvature changes abruptly. Based on these findings and the comparison with figures 8 and 9, it is evident that the curvature of the path significantly impacts the accuracy of interpolation, with smoother motion observed for paths with lower curvature and minimal changes in curvature.

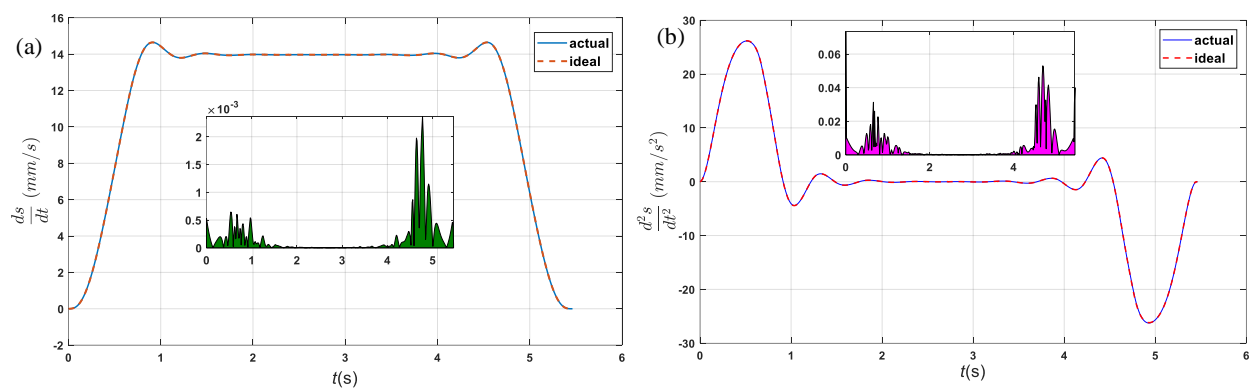

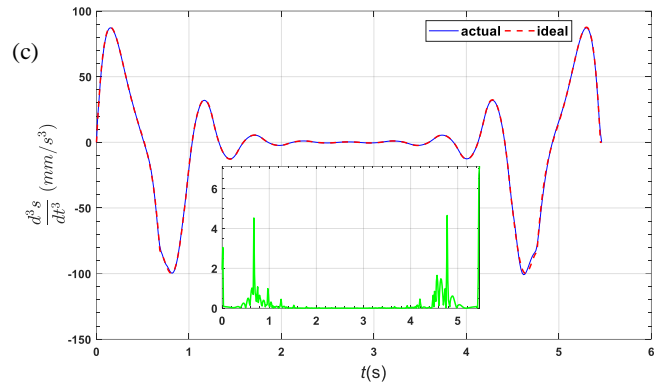

**Fig.14.** comparative analysis of motion profiles for the B-spline path displayed in Figure 17, between the ideal and actual states. The depicted curves represent: **(a)** velocity, **(b)** acceleration, and **(c)** jerk.

By utilizing the method described in subsection 3.3, the spatial orientation of the tool was represented using quaternions. These quaternions were fitted to the B-spline curve and parameterized with the parameter $w$. The parameter $w$ was further reparametrized using Bezier curves, which aimed to minimize changes in curvature compared to the path length parameter, $s$. Additionally, the path length was parameterized using trajectory generation method from section 4. Consequently, in each interpolation period, the quaternions representing the spatial orientation of the tool were obtained. Figure 15, part (a), illustrates these quaternions after omitting one dimension on the unit sphere. Part (b) of Figure 15 presents the variations of the real and imaginary components of the quaternions in relation to the curve parameter $w$. It can be observed that due to the mechanism's rotational degree of freedom around x-axis, the changes in the imaginary parts $j$ and $k$ are zero. Furthermore, the shape of the curve representing the variations in the quaternions closely resembles the graph (refer to Figure 8) depicting the feed-rate of the path length with respect to time. The accuracy of parametrizing the nonlinear relationship between $s$ and $w$ is evident in part (b) of Figure 15. Also, Figure 16 demonstrates how the curve parameter $w$ is effectively fitted using piece-wise continuous Bezier curves with high precision.

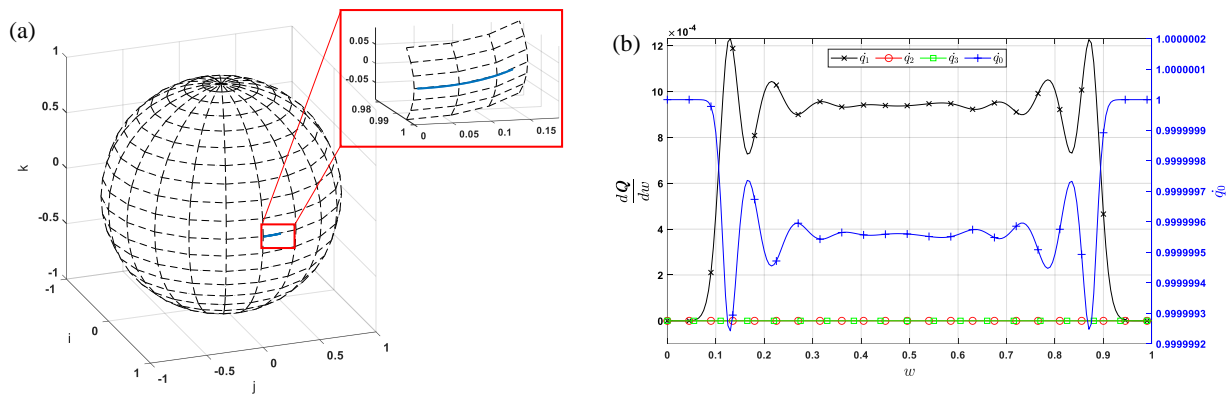

**Fig.15.** Visualization of tool's angular orientation quaternions: **(a)** Representation of the curve on the unit sphere, derived from fitting B-spline curves on the purely imaginary quaternions. **(b)** Variation of both imaginary and real components of the unit quaternions with respect to the curve parameter, $w$.

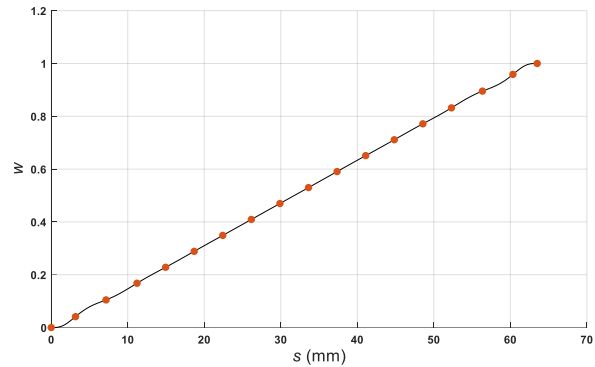

**Fig.16.** Correlation between path length and curve parameter of tool orientation quaternions achieved through piece-wise Bezier curves optimization algorithm.

Furthermore, Figure 17 illustrates the angular velocity, angular acceleration, and angular jerk graphs resulting from encoder feedback in comparison to their ideal counterparts. The minimal deviations between the values obtained from direct kinematics and the ideal values are evident in all three parts of the figure. Specifically, from part (a) of Figure 22, the maximum deviation of angular velocity is 1.2198e-5 (deg/s) with an average of 2.3606e-7 (deg/s) along the path; from part (b) of Figure 22, the maximum deviation of angular acceleration is 8.969e-4 (deg/s²) with an average of 1.688e-5 (deg/s²) along the path; and from part (c) of Figure 17, the maximum angular jerk deviation is 0.05 (deg/s³) with an average of 0.001 (deg/s³) along the path.

An interesting observation from Figure 17 is the striking resemblance between the motion curves of the angular orientation of the tool and the tool position depicted in Figures 8 and 9. As previously mentioned in section 3, considering the absence of an analytical relationship between tool orientation and path length, the close correspondence of the motion curves obtained from direct kinematics with the ideal values highlights the success of the parametric interpolation method for the tool's spatial orientation, as described in this article. The results obtained through this method prove to be favorable and reliable.

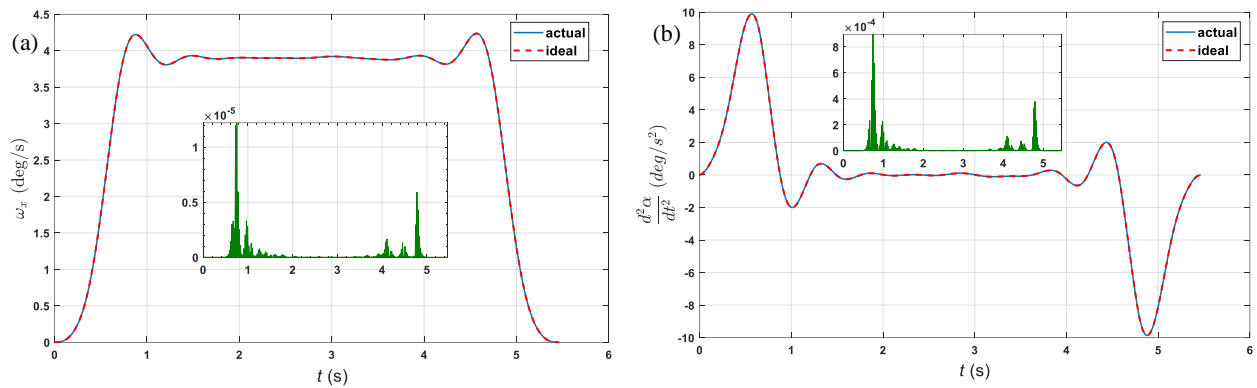

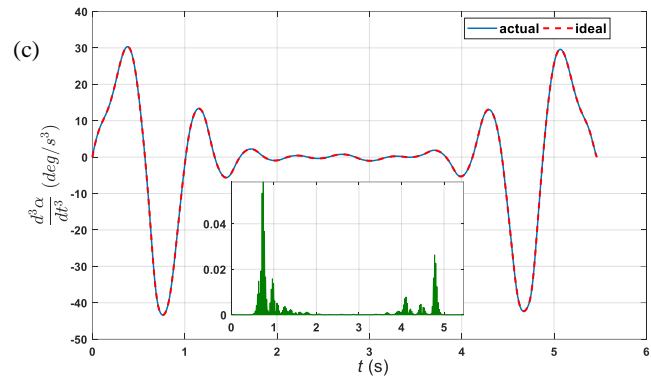

**Fig.17.** Comparison of angular motion curves in the ideal state and the results of encoder data and direct kinematics: (a) Depicts angular velocity, (b) shows angular acceleration, and (c) presents angular jerk.

## 6. Conclusion

This study introduced and implemented a dual-stage smooth trajectory and hybrid B-spline-Quaternion interpolation method for freeform paths in a four-degree-of-freedom multi-axis parallel kinematic milling robot. The research commenced by presenting the mechanical structure and establishing kinematic relations between the actuator and task space. Subsequently, trajectory interpolation and generation for B-spline tool paths were outlined, and a comprehensive comparison with existing interpolation methods from the literature was conducted. Moreover, position interpolation utilized modifier polynomials, while angular orientation interpolation employed unit quaternions. The synchronization of path length and spatial orientation was achieved through an optimization process employing piece-wise Bezier curves, minimizing the integral norm of geometric jerk. Further, motion planning involved solving an optimization problem for minimum jerk and time-optimal Bezier curve minimization, with the two-layer approach encompassing both task and actuator spaces, and implemented within an arm-based embedded system. Experimental validation entailed writing and executing all interpolation and motion planning steps within a C++ object-oriented programming environment, with GUI development facilitated by the Qt framework and multithreaded Ethernet communication with hardware.

In this research, several key conclusions have been drawn. Natural interpolation methods are unsuitable for highly curved or rapidly changing paths, leading to undesirable velocity oscillations, pronounced acceleration fluctuations, and uncontrollable jerk. The first-order Taylor expansion method accumulates rounding errors and struggles with sharp corners, contributing to computational inefficiency. The second-order Taylor expansion method, while better in approximating velocity, is sensitive to cumulative rounding errors and less practical for real-time applications. In contrast, modifier polynomials provide high-quality interpolation with reduced velocity fluctuations, enhanced smoothness, and computational advantages. A novel synchronization method for path length and tool orientation quaternions ensures accurate results. Overall, the developed methodology demonstrates remarkable efficiency and performance improvements in multi-axis tool path interpolation for robotic systems. By reducing velocity fluctuations and enhancing smoothness through modifier polynomials, it ensures more precise and controlled movements, particularly beneficial in applications requiring high accuracy, such as milling machines. Moreover, the synchronization method for path length and tool orientation quaternions enhances overall performance, resulting in close correspondence between actual and ideal values for velocity, acceleration, and angular jerk. This methodology's versatility and adaptability make it a valuable asset for subtractive and additive manufacturing with robots, offering potential applications across machining processes, where precise and efficient motion planning is paramount.

**Declaration of competing interest**

The authors declare that they have no known competing financial interests or personal relationships that could have appeared to influence the work reported in this paper.

## Acknowledgments

None to report.